\documentclass[10pt,twocolumn,letterpaper]{article}

\usepackage{cvpr}
\usepackage{times}
\usepackage{epsfig}
\usepackage{graphicx}
\usepackage{amsmath}
\usepackage{amssymb}
\usepackage{multirow}
\usepackage{subfigure}
\usepackage[pagebackref=true,breaklinks=true,letterpaper=true,colorlinks,bookmarks=false]{hyperref}

\cvprfinalcopy 

\def\cvprPaperID{1799} 

\ifcvprfinal\fi
\begin{document}

\title{Action Recognition with Trajectory-Pooled Deep-Convolutional Descriptors}

\author{Limin Wang$^{1,2}$ \quad \quad Yu Qiao$^{2}$ \quad \quad Xiaoou Tang$^{1,2}$ \\
\small $^{1}$Department of Information Engineering, The Chinese University of Hong Kong\\
\small $^{2}$Shenzhen key lab of Comp. Vis. \& Pat. Rec.,  Shenzhen Institutes of Advanced Technology, CAS, China \\
{\tt\small 07wanglimin@gmail.com, yu.qiao@siat.ac.cn, xtang@ie.cuhk.edu.hk}
}

\maketitle

\begin{abstract}
Visual features are of vital importance for human action understanding in videos. This paper presents a new video representation, called \emph{trajectory-pooled deep-convolutional descriptor (TDD)}, which shares the merits of both hand-crafted features \cite{WangS13a} and deep-learned features \cite{SimonyanZ14}. Specifically, we utilize deep architectures to learn discriminative convolutional feature maps, and conduct trajectory-constrained pooling to aggregate these convolutional features into effective descriptors. To enhance the robustness of TDDs, we design two normalization methods to transform convolutional feature maps, namely spatiotemporal normalization and channel normalization. The advantages of our features come from (i) TDDs are automatically learned and contain high discriminative capacity compared with those hand-crafted features; (ii) TDDs take account of the intrinsic characteristics of temporal dimension and introduce the strategies of trajectory-constrained sampling and pooling for aggregating deep-learned features. We conduct experiments on two challenging datasets: HMDB51 and UCF101. Experimental results show that TDDs outperform previous hand-crafted features \cite{WangS13a} and deep-learned features \cite{SimonyanZ14}. Our method also achieves superior performance to the state of the art on these datasets \footnote{The TDD code and learned two-stream ConvNet models are available at \url{https://wanglimin.github.io/tdd/index.html}}.
\end{abstract}

\section{Introduction}
\begin{figure}
  \includegraphics[width=\linewidth]{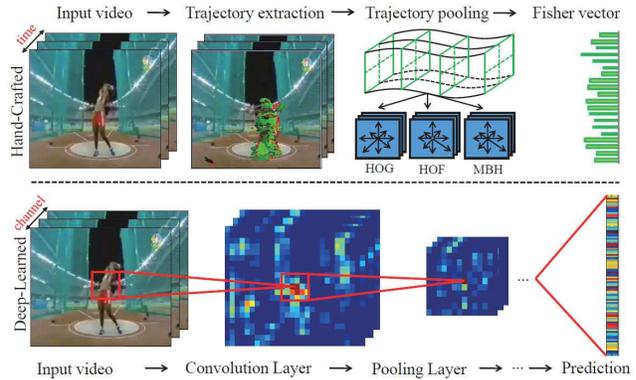}
  \caption{There are mainly two types of features in action recognition: \emph{hand-crafted} features and \emph{deep-learned} features. For hand-crafted features, improved trajectories \cite{WangS13a} combined with Fisher vector are most successful. For deep-learned features, Convolutional Networks (ConvNets) \cite{lecun-01a} are popular deep architectures, which contain a sequence of convolutional and pooling layers. They aims to automatically learn features with a deep discriminatively trained neural network. }
  \label{fig:motivation}
  \vspace{-5mm}
\end{figure}

Human action recognition \cite{AggarwalR11,SimonyanZ14,WangS13a,WangQT13a,WangQT14a} in videos attracts increasing research interests in computer vision community due to its potential applications in video surveillance, human computer interaction, and video content analysis. However, action recognition remains as a difficult problem when focusing on realistic datasets collected from movies \cite{LaptevMSR08}, web videos \cite{KuehneJGPS11,Soomro12}, and TV shows \cite{Patron-PerezMRZ12}. There are large intra-class variations in the same action class, which may be caused by background clutter, viewpoint change, and various motion speeds and styles. Meanwhile, the high dimension and low resolution of video further increases the difficulty to design efficient and robust recognition method. \emph{Visual representations} from action videos are crucial for dealing with these issues and designing effective recognition systems. Currently, there are mainly two types of video features available for action recognition, as illustrated in Figure \ref{fig:motivation}.

The first type of representations are the \emph{hand-crafted} local features, and typical local features include Space Time Interest Points \cite{Laptev05}, Cuboids \cite{Dollar05}, Dense Trajectories \cite{WangKSL13}, and Improved Trajectories \cite{WangS13a}. Calculation of these local features can be usually decomposed into two phrases: \emph{detector}, which aims to discover the salient and informative regions for action understanding, and \emph{descriptor}, whose goal is to describe the visual patterns of extracted regions. Among these local features, improved trajectories with rich descriptors of HOG, HOF, MBH have shown to be successful on a number of challenging datasets (e.g. HMDB51 \cite{KuehneJGPS11}, UCF101 \cite{Soomro12}) and contests (e.g. THUMOS \cite{THUMOS13}). Improved trajectories include several important ingredients in their extraction process. Firstly, these extracted trajectories are mainly located at regions with high motion salience, which contain rich and discriminative information for action recognition. Secondly, these local descriptors of the corresponding regions in several successive frames, are aligned and pooled along the trajectories. This trajectory-constrained sampling strategy also takes account of the temporal continuity of human action, and is effective to deal with the variations of motion speed. However, these hand-crafted descriptors are not optimized for visual representation and may lack discriminative capacity for action recognition.

The second type of representations are the \emph{deep-learned} features, and typical methods include Convolutional RBMs \cite{TaylorFLB10}, 3D ConvNets \cite{JiXYY13}, Deep ConvNets \cite{KarpathyTSLSF14}, and Two-Stream ConvNets \cite{SimonyanZ14}. These deep learning methods aim to automatically learn the semantic representation from raw video by using a deep neural network discriminatively trained from a large number of labeled data. Two-Stream ConvNets \cite{SimonyanZ14} are probably the most successful architecture at present, and they match the state-of-the-art performance of improved trajectories \cite{WangS13a,WangS13b} on UCF101 and HMDB51. They are composed of two neural networks, namely spatial nets and temporal nets. Spatial nets mainly capture the discriminative appearance features for action understanding, while temporal nets aim to learn the effective motion features. However, unlike image classification tasks \cite{KrizhevskySH12}, these deep learning based methods fail to outperform previous hand-crafted features. One problem of deep learning methods is that they require a large number of labeled videos for training, while most available datasets are relatively small. Meanwhile, most of current deep learning based action recognition methods largely ignore the intrinsic difference between temporal domain and spatial domain, and just treat temporal dimension as feature channels when adapting the architectures of ConvNets to model videos.

Motivated by the above analysis, this paper proposes a new kind of video feature, called \emph{trajectory-pooled deep-convolutional descriptor} (TDD). The design of TDD aims to combine the benefits of both hand-crafted and deep-learned features. To achieve this goal, our approach integrates the key factors from two successful video representations, namely improved trajectories \cite{WangS13a} and two-stream ConvNets \cite{SimonyanZ14}. We utilize deep architecture to learn multi-scale convolutional feature maps, and introduce the strategies of trajectory-constrained sampling and pooling to encode deep features into effective descriptors.

Specifically, we first train two-stream ConvNets on a relatively large dataset, while more labeled action videos will make ConvNet training more stable and robust. Then, we treat the learned two-stream ConvNets as generic feature extractors, and use them to obtain multi-scale convolutional feature maps for each video. Meanwhile, we detect a set of point trajectories with the method of improved trajectories. Based on convolutional feature maps and improved trajectories, we pool the local ConvNet responses over the spatiotemporal tubes centered at the trajectories, where the resulting descriptor is called TDD. Finally, we choose Fisher vector representation to aggregate these local TDDs over the whole video into a global super vector, and use linear SVM as the classifier to perform action recognition. We conduct experiments on two public action datasets: the HMDB51 dataset \cite{KuehneJGPS11} and the UCF101 dataset \cite{Soomro12}. We show that our TDDs obtain the state-of-the-art performance for action recognition on these challenging datasets. Meanwhile, our results demonstrate that our TDDs are complementary to those hand-crafted features (HOG, HOF, and MBH) and the fusion of them is able to further boost the recognition performance.

\section{Related Works}
\begin{figure*}[t]
  \includegraphics[width=\textwidth]{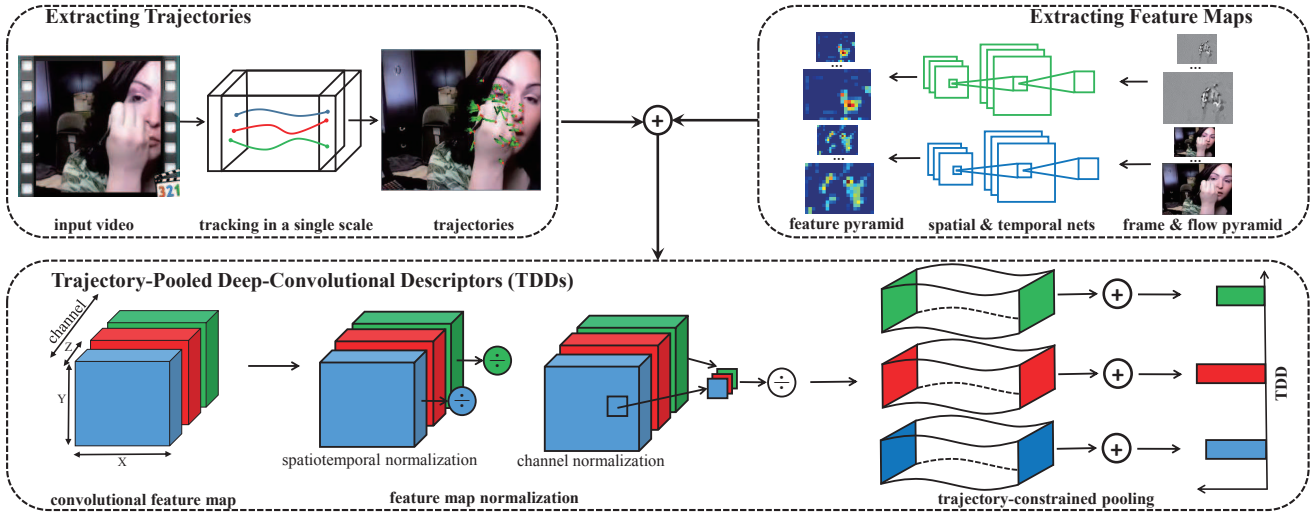}
  \caption{\textbf{Pipeline of TDD.} The whole process of extracting TDD is composed of three steps: (i) extracting trajectories, (ii) extracting multi-scale convolutional feature maps, and (iii) calculating TDD. We effectively exploit two available state-of-the-art video representations, namely improved trajectories and two-stream ConvNets. Grounded on them, we conduct trajectory-constrained sampling and pooling over convolutional feature maps to obtain trajectory-pooled deep convolutional descriptors.}
  \label{fig:pipeline}
  \vspace{-5mm}
\end{figure*}

\textbf{Hand-crafted features.} Local features \cite{Dollar05,Laptev05,WangUKLS09,WillemsTG08} have become popular and effective representations in action recognition, as these local features do not require algorithms to detect human body and are robust to background clutter, illumination changes, and video noise. Space Time Interest Points \cite{Laptev05} proposed Harris3D detector to extract informative regions, while Cuboid \cite{Dollar05} detector relied on temporal Gabor filters. Willems \emph{et al.} \cite{WillemsTG08} proposed a Hessian detector, which is a spatio-temporal extension of Hessian saliency measure used for blob detection in images. Meanwhile several local descriptors have been proposed to represent the 3D volumes extracted around these interest points, such as Histogram of Gradient (HOG), Histogram of Optical Flow (HOF) \cite{LaptevMSR08}, 3D Histogram of Gradient (HOG3D) \cite{KlaserMS08}, and Extended SURF (ESURF) \cite{WillemsTG08}. Recent works made use of point trajectories \cite{WangKSL13,WangS13a} to extract and align 3D volumes, and resorted to more rich low level descriptors for constructing effective video representations, including HOG, HOF, and Motion Boundary Histogram (MBH).

One limitation of these local features is that they lack semantics and discriminative capacity. To overcome this issue, several mid-level and high-level video representations have been proposed such as Action Bank \cite{SadanandC12}, Dynamic-Poselets \cite{WangQT14b}, Motionlets \cite{WangQT13a}, Motion Atoms and Phrases \cite{WangQT13b}, and Actons \cite{ZhuWYZT13}. They usually resorted to some heuristic mining methods to select discriminative visual elements as feature units. Instead, this paper takes a different view of this problem and replace these local hand-crafted descriptors with deep-learned representations. Our deep representations deliver high level semantic information, and are learned automatically from training data without using these heuristic rules.

\textbf{Deep-learned features.} Deep learning techniques have achieved great success in image based tasks \cite{KrizhevskySH12,SimonyanZ14a,SzegedyLJSRAEVR14,ZeilerF14} and there have been a number of attempts to develop deep architectures for video action recognition \cite{JiXYY13,KarpathyTSLSF14,SimonyanZ14,TaylorFLB10}. Taylor \emph{et al.} \cite{TaylorFLB10} used Gated Restricted Boltzmann Machines (GRBMs) to learn the motion features in an unsupervised manner and then resorted to convolutional learning to fine tune the parameters. Ji \emph{et al.} \cite{JiXYY13} extended 2D ConvNet to video domain for action recognition on relatively small datasets, and recently Karpathy \emph{et al.} \cite{KarpathyTSLSF14} tested ConvNets with deep structures on a large dataset, called Sports-1M. However, these deep models achieved lower performance compared with shallow hand-crafted representation \cite{WangS13a}, which might be ascribed to two facts: firstly, available action datasets are relatively small for deep learning; secondly, learning complex motion patterns is more challenging. Simonyan \emph{et al.} \cite{SimonyanZ14} designed two-stream ConvNets containing spatial and temporal net by exploiting large ImageNet dataset for pre-training and explicitly calculating optical flow for capturing motion information, and finally it matched the state-of-the-art performance.

However, these deep models lacked considerations of temporal characteristics of video data and relied on large training datasets. We incorporate video temporal characteristics into deep architectures by using strategy of trajectory-constrained sampling and pooling, and propose a new descriptor. Meanwhile, our descriptors can be easily adapted to the datasets of smaller size.

\section{Improved Trajectories Revisited}
\label{sec:idt}
As shown in Figure \ref{fig:pipeline}, our proposed representation (TDD) is based on low level trajectory extraction and we choose improved trajectories \cite{WangS13a}. In this section, we briefly review the extraction process of improved trajectories. It is worth noting that our TDD is independent of the method of extracting trajectories, and we use improved trajectories due to its good performance.

Improved trajectories are extended from dense trajectories \cite{WangKSL13}. To compute dense trajectories, the first step is to densely sample a set of points on 8 spatial scales on a grid with step size of 5 pixels. Points in homogeneous areas are eliminated by setting a threshold for the smaller eigenvalue of their autocorrelation matrices. Then these sampled points are tracked by media filtering of dense flow field.
\begin{equation}
  P_{t+1} = (x_{t+1},y_{t+1}) = (x_t,y_t) + (\mathcal{M} * \omega_t) |_{(\overline{x}_t,\overline{y}_t)},
\end{equation}
where $\mathcal{M}$ is the median filter kernel, $*$ is convolutional operation, $\omega_t=(u_t,v_t)$ is the dense optical flow field of the $t^{th}$ frame, and $(\overline{x}_t,\overline{y}_t)$ is the rounded position of $(x_t,y_t)$. To avoid the drifting problem of tracking, the maximum length of trajectory is set as 15-frame. Finally, those static trajectories are removed as they lack motion information, and other trajectories with suddenly large displacement are also ignored, since they are obviously incorrect due to inaccurate optical flow.

Improved trajectories boost the recognition performance of dense trajectories by taking camera motion into account. It assumes that the background motion of two consecutive frames can be characterized by a homography matrix. To estimate the homography matrix, the first step is to find the correspondence between two consecutive frames. They resort to SURF \cite{BayTG06} feature matching and optical flow based matching, as these two kinds of matching scheme are complementary to each other. Then, they use the RANSAC \cite{FischlerB81} algorithm to estimate homography matrix. Based on the homography, they rectify the frame image to remove the camera motion and re-calculate the optical flow, called \emph{warped flow}. Warped flow brings advantages to the descriptors calculated from optical flows, in particular for HOF, and trajectories corresponding to camera motion can be removed too.

We adopt improved trajectories for the task of TDD extraction, but make a modification. Unlike dense trajectories or improved trajectories, we only track points on its original spatial scale, and extract multi-scale TDDs around the extracted trajectories (see Section \ref{sec:dcp}). We observe that tracking on a single scale is fast for implementation. In summary, given a video $V$, we obtain a set of trajectories
\begin{equation}
  \mathbb{T}(V) = \{T_1,T_2,\cdots,T_K\},
\end{equation}
where $K$ is the number of trajectories, and $T_k$ denotes the $k^{th}$ trajectory in the original spatial scale:
\begin{equation}
  T_k =\{(x^k_1,y^k_1,z^k_1),(x^k_2,y^k_2,z^k_2),\cdots,(x^k_P,y^k_P,z^k_P)\},
  \label{equ:tra}
\end{equation}
where $(x^k_p,y^k_p,z^k_p)$ is the pixel position of the $p^{th}$ point in trajectory $T_k$, and $P$ is the length of trajectory ($P=15$). These trajectories will be used for trajectory-constrained sampling and pooling in the process of TDD extraction, as described in the next section.

\section{Deep Convolutional Descriptors}
\label{sec:dcp}
In this section, we describe a new video representation, called \emph{trajectory-pooled deep-convolutional descriptor} (TDD), which shares the benefits of both hand-crafted and deep-learned features. We first introduce the architectures of convolutional networks (ConvNets) we used. Then, we show how to adapt the ConvNets trained on large datasets to extract multi-scale convolutional feature maps. Finally, based on improved trajectories and convolutional feature maps, we describe the details of how to calculate TDDs.

\subsection{Convolutional networks}
\begin{table*}
\centering
\small
\resizebox{\textwidth}{!}{
\begin{tabular}{|c|c|c|c|c|c|c|c|c|c|c|c|}
  \hline
  Layer & conv1  & pool1 & conv2 & pool2 & conv3 & conv4 & conv5 & pool5 & full6 & full7 & full8 \\
  \hline
  \hline
  size & $7 \times 7 $ & $3 \times 3 $& $5 \times 5$ & $3 \times 3 $ & $3\times3$  & $3\times3$  &$3\times3$   & $3\times3$  & - & - & - \\
  stride & 2 & 2 & 2 & 2 & 1 & 1 & 1 & 2 & - & - & - \\
  channel & 96 & 96 & 256 & 256 & 512  & 512 & 512 & 512 & 4096 & 2048 & 101 \\
  \hline
  \hline
  map size ratio & 1/2 & 1/4 & 1/8 & 1/16 & 1/16 & 1/16 & 1/16 & 1/32 & - & - & - \\
  receptive field & $7 \times 7 $  & $11 \times 11$ & $27 \times 27$ & $43 \times 43$ & $75 \times 75$ & $107 \times 107$ & $139 \times 139 $& $171 \times 171 $ & -  & - &  - \\
  \hline
\end{tabular}
}
\vspace{1mm}
\caption{\textbf{ConvNet Architectures.} We use similar architectures to two-stream ConvNets \cite{SimonyanZ14}, which are adapted to the task of action recognition from the Clarifai networks \cite{ZeilerF14}, with less filters in conv4 layer (512 vs. 1024) and lower-dimensional full7 layer (2048 vs. 4096). For layers of conv1 and conv2, local response normalized (LRN) is applied with parameters settings: $n =5, \alpha=5 \times 10^{-4}, \beta=0.75$. The layers of full6 and full7 are regularised by using dropout and the full8 layer acts as a soft-max classifier. The activation function for all weight layers is the rectification linear unit (RELU). The size ratios of feature maps with respect to input data range from $1/2$ to $1/32$, and the feature receptive fields vary from $7 \times 7$ to $171 \times 171$, for different convolutional and pooling layers (conv1 to pool5).}
\label{tbl:convnet}
\vspace{-5mm}
\end{table*}

Our TDD starts with designing deep ConvNets for extracting convolutional feature maps. In principle, any kind of ConvNet architecture can be adopted for TDD extraction. In our implementation, we choose the two-stream ConvNets\cite{SimonyanZ14} due to their good performance on the datasets of UCF101 and HMDB51.

The two-stream ConvNets contain two separate ConvNets, namely spatial nets and temporal nets. Spatial nets are designed for capturing static appearance cues, which are trained on single frame images ($224 \times 224 \times 3$), while temporal nets aim to describe the dynamic motion information, whose input are volumes of stacking optical flow fields ($224 \times 224 \times \ 2F$, $F$ is the number of stacking flows). Meanwhile, decoupling the spatial and temporal nets also allows to exploit the available images by pre-training spatial nets on the ImageNet challenge dataset \cite{DengDSLL009}, and explicitly handle motion information with optical flow algorithms for temporal nets

The details about ConvNets are shown in Table \ref{tbl:convnet}. This ConvNet architecture is original from the Clarifai networks \cite{ZeilerF14} and adapted to the task of action recognition with less filters in conv4 layer and lower-dimensional full7 layer. But we make a small modification. We use the same network architecture for both spatial and temporal net in addition to the input data layer, while the original two-stream ConvNets \cite{SimonyanZ14} ignore the second local response normalized (LRN) layer in the temporal net due to memory consumption problem. The implementation and training details can be found in Section \ref{sec:exp}.

\subsection{Convolutional feature maps}
Once the training of two-stream ConvNets is complete, we treat them as generic feature extractors to obtain the convolutional feature maps of videos. In general, for each video, we obtain these feature maps of spatial and temporal net in a frame-by-frame and volume-by-volume manner, respectively. In order to make the feature maps with equal temporal duration with input video, we pad the optical flow fields at the beginning with $F-1$ copies of the optical flow field from the first frame, where $F$ is the number of stacking optical flow.

For each frame or volume, we take it as the input for spatial or temporal nets. We make two modifications about the spatial and temporal nets. The first one is that we remove the layers after the target layer for feature extraction. For example, to extract feature maps of conv4, we will remove the layers from conv5 to full8. Therefore, the output of spatial and temporal net will be the convolutional feature maps, which will be used for extracting TDD in the next subsection.

The second modification is that before each convolutional or pooling layer, with kernel size $k$, we conduct zero padding of the layer's input with size $\lfloor k/2\rfloor$. This padding allows the input and output maps of these layers to have the same spatial extent. With this padding, it will be straightforward to map the positions of trajectory points in
video to the coordinates of convolutional feature maps. A trajectory point with video coordinates  $(x_p,y_p,z_p)$ in Equation (\ref{equ:tra}) will be centered on $(r \times x_p, r \times y_p, z_p)$ in convolutional map, where $r$ is map size ratio with respective to input size, as listed in Table \ref{tbl:convnet}.

ConvNets are bottom-up architectures with a sequence of alternating convolutional and pooling layers. Different layers of ConvNets have various receptive fields as shown in Table \ref{tbl:convnet}, ranging from $7 \times 7$ to $171 \times 171$. As described in paper \cite{ZeilerF14}, these different layers capture patterns from simple visual elements such as edges, to complex visual concepts such as parts and objects. The higher layers have larger receptive fields and obtain more invariant and discriminative features. Intuitively, these different layers describe the visual content at different levels, each of which is complementary to each other for the task of recognition. We will exploit this complimentary property of different layers during the extraction of TDD. Given a video $V$, we obtain a set of convolutional feature maps:
\begin{equation}
  \mathbb{C}(V) = \{C^s_1,C^s_2,\cdots,C^s_M,C^t_1,C^t_2,\cdots,C^t_M\},
  \label{equ:map}
\end{equation}
where $C^s_m \in \mathbb{R}^{H_m \times W_m \times L \times N_m}$ is the $m^{th}$ feature map of spatial net, $H_m$ is its height, $W_m$ is its width, $L$ is the video duration, and $N_m$ is the number of channels. $C^t_m \in \mathbb{R}^{H_m \times W_m \times L \times N_m} $ is the $m^{th}$ feature map of temporal net, $M$ is the number of layers for extracting TDD.

\subsection{Trajectory-pooled descriptors}
\label{sec:TDD}
We will describe the method for extracting trajectory-pooled deep-convolutional descriptors (TDDs) from a set of improved trajectories $\mathbb{T}(V)$ and convolutional feature maps $\mathbb{C}(V)$ for a given video $V$. In essence, TDD is a kind of local trajectory-aligned descriptor computed in a 3D volume around the trajectory. TDDs from the spatial and temporal nets capture the appearance and motion information of this 3D volume, respectively. The size of the volume is $N \times N$ pixels and $P$ frames, where $N$ is the receptive field size and $P$ is the trajectory length. The extraction of TDD is composed of two steps: \emph{feature map normalization} and \emph{trajectory pooling}.

Normalization proves to be an effective strategy in designing features partially because it can reduce the influence of illumination. It has been widely exploited in local descriptors such as SIFT \cite{Lowe04}, HOG \cite{DalalT05}, and HOF \cite{LaptevMSR08}, and in deep learning such as local response normalization \cite{KrizhevskySH12}. We apply the normalization strategy to the convolutional feature maps of two-stream ConvNets to suppress the activation burstiness of some neurons. We design two kinds of normalization methods:
\begin{itemize}
  \item \emph{Spatiotemporal Normalization.} For spatiotemporal normalization, we normalize the feature map for each channel independently across the video spatiotemporal extent. Given a feature map $C \in \mathbb{R}^{H \times W \times L \times N}$ of Equation (\ref{equ:map}), we normalize the convolutional feature value as follows:
      \begin{equation}
        \widetilde{C}_{st}(x,y,z,n) = C(x,y,z,n) / \mathrm{maxV}_{st}^n,
      \end{equation}
      where $\mathrm{maxV}_{st}^n$ is the maximum value of $n^{th}$ feature maps over the whole video spatiotemporal extent, which means $\mathrm{maxV}_{st}^n = \max_{x,y,z} C(x,y,z,n)$. The spatiotemporal normalization method ensures that each convolutional feature channel ranges in the same interval, and thus contributes equally to final TDD recognition performance.
  \item \emph{Channel Normalization.} For channel normalization, we normalize the feature map for each pixel independently across the feature channels. We conduct channel normalization for feature map $C \in \mathbb{R}^{H \times W \times L \times N}$ as follows:
      \begin{equation}
        \widetilde{C}_{ch}(x,y,z,n) = C(x,y,z,n) / \mathrm{maxV}_{ch}^{x,y,z},
      \end{equation}
      where $\mathrm{maxV}_{ch}^{x,y,z}$ is the maximum value of different feature channels at pixel position $(x,y,z)$, that is $\mathrm{maxV}_{ch}^{x,y,z} = \max_{n} C(x,y,z,n)$. This channel normalization is able to make sure that the feature value of each pixel range in the same interval, and let each pixel make the equal contribution in the final representation.
\end{itemize}

After the step of feature normalization, we will extract TDDs based on trajectories and normalized convolutional feature maps by using trajectory pooling. Specifically, given a trajectory $T_k$ and a normalized feature map $\widetilde{C}_m^{a}$, which is the $m^{th}$-layer feature map after either spatiotemporal normalization or channel normalization from spatial net or temporal net ($a \in \{s,t\}$), we conduct sum-pooling of the normalized feature maps over the 3D volume centered at the trajectory as follows:
\begin{equation}
  D(T_k,\widetilde{C}_m^{a}) = \sum_{p=1}^P \widetilde{C}_m^{a}(\overline{(r_m \times x_p^k)},\overline{(r_m \times y_p^k)},z_p^k),
  \label{equ:TDD}
\end{equation}
where $(x_p^k,y_p^k,z_p^k)$ is the $p^{th}$ point position of video coordinates in trajectory $T_k$, $r_m$ is the $m^{th}$-layer map size ratio with respective to input size as listed in Table \ref{tbl:convnet}, $\overline{(\cdot)}$ is the rounding operation. $D(T_k,\widetilde{C}_m^{a})$ is called \emph{trajectory-pooled deep convolutional descriptor}, and is a new kind of feature combing the merits of both improved dense trajectories and two-stream ConvNets.

\textbf{Multi-scale TDD extension.} The above description on TDD extraction is about the single scale, we will present the multi-scale extension of TDD. For improved trajectory, it samples points and tracks them on multi-scale videos, while fixes the spatial extent of HOG, HOF, and MBH descriptors as $32 \times 32$. The original method needs to conduct point tracking and descriptor calculation in multi-scale settings. In our implementation, we try a more efficient multi-scale strategy. Specifically, we calculate optical flow and track point in a single scale. Then we construct multi-scale pyramid representations of video frames and optical flow fields. These pyramid representations are fed into the two stream ConvNets and transformed into multi-scale convolutional feature maps as shown in Figure \ref{fig:pipeline}. Based on multi-scale convolutional maps and single-scale improved trajectories, we are able to compute multi-scale TDDs efficiently, by applying trajectory pooling to multi-scale convolutional feature maps as described above. The only modification to different scales is to replace feature map size ratio $r_m$ in Equation (\ref{equ:TDD}) with $r_m \times s$, where $s$ is the scale of current feature map. In practice, compared with improved trajectories, we use less scales with $s = 1/2, 1/\sqrt{2},1,\sqrt{2},2$.

\section{Experiments}
\label{sec:exp}
In this section, we first present the details of datasets and their evaluation scheme. Then, we describe the details of our method. Finally, we give the experimental results and compare TDD with the state of the art.

\subsection{Datasets}
In order to verify the effectiveness of TDDs, we conduct experiments on two public large datasets, namely HMDB51 \cite{KuehneJGPS11} and UCF101 \cite{Soomro12}. The HMDB51 dataset is a large collection of realistic videos from various sources, including movies and web videos. The dataset is composed of $6,766$ video clips from $51$ action categories, with each category containing at least 100 clips. Our experiments follow the original evaluation scheme using three different training/testing splits. In each split, each action class has $70$ clips for training and $30$ clips for testing. The average accuracy over these three splits is used to measure the final performance.

The UCF101 dataset contains 101 action classes and there are at least 100 video clips for each class. The whole dataset contains $13,320$ video clips, which are divided into 25 groups for each action category. We follow the evaluation scheme of the THUMOS13 challenge \cite{THUMOS13} and adopt the three training/testing splits for evaluation. As UCF101 is larger than HMDB51, we use the UCF101 dataset to train two-stream ConvNets initially, and transfer this learned model for TDD extraction on the HMDB51 dataset.

\subsection{Implementation details}
\textbf{Two-stream ConvNets training.} Training deep ConvNets is more challenging for action recognition as action is more complex than object and the available dataset is extremely small compared with the ImageNet dataset \cite{DengDSLL009}. We choose the training dataset of UCF101 split1 for learning two-stream ConvNets as it is probably the largest public available dataset. We use the Caffe toolbox \cite{JiaSDKLGGD14} for ConvNet implementation. The network weights are learnt using the mini-batch (set to 256) stochastic gradient descent with momentum (set to 0.9). For spatial net, we first resize the frame to make the smaller side as $256$, and then randomly crop a $224 \times 224$ region from the frame. It then undergoes random horizontal flipping. We pre-train the network with the public available model \cite{Chatfield14}. Finally, we fine tune the model parameters on the UCF101 dataset, where the learning rate is set as $10^{-2}$, decreased to $10^{-3}$ after 14K iterations, and training stopped at $20K$ iterations.

For temporal net, its input is 3D volume of stacking optical flows fields. We choose the TVL1 optical flow algorithm \cite{ZachPB07} and use the OpenCV implementation, due to its balance between accuracy and efficiency. For fast computation, we discretize the values of optical flow fields into integers and set their range as $0$-$255$ just like images. Specifically, we choose to stack 10 frames of optical flow fields to keep a balance between performance and efficiency. We train temporal net on UCF101 from scratch. As the dataset is relatively small, we use high dropout ratio to improve the generalization capacity of trained model. We set dropout 0.9 for full6 layer and dropout 0.8 for full7 layer. The training procedure of temporal net is similar to spatial net and a $224 \times 224 \times 20$ sub-volume is randomly cropped and flipped from training video. The learning rate is initially set as $10^{-2}$ and decreases to $10^{-3}$ after 50K iterations. It is then reduced to $10^{-4}$ after 70K iterations and training is stopped at 90K iterations.

\textbf{Results of two-stream ConvNets.} To evaluate the trained model, as in \cite{SimonyanZ14}, we select $25$ frames for each video clip and obtain 10 crops for each frame. The final recognition result is the average across these crops and frames. We obtain $71.2\%$ recognition accuracy with spatial net and $80.1\%$ with temporal net. The performance of our implemented two-stream ConvNets is $84.7\%$, which is similar to that of two-stream ConvNets \cite{SimonyanZ14} ($85.6\%$). However, obtaining ConvNets with high performance is not the final goal of this paper, and we aim to verify the effectiveness of TDDs.

\textbf{Feature encoding.} We choose Fisher vector \cite{SanchezPMV13} to encode the TDDs of a video clip into high dimensional representation as its effectiveness for action recognition has been verified in previous works \cite{WangWQ12,SunN13}, and then use a linear SVM as the classifer ($C=100$). In order to train GMMs, we first de-correlate TDD with PCA and reduce its dimension to $D$. Then, we train a GMM with $K$ ($K=256$) mixtures, and finally the video is represented with a $2KD$-dimensional vector.

\subsection{Exploration experiments}
\begin{figure}
  \includegraphics[width=0.50\linewidth]{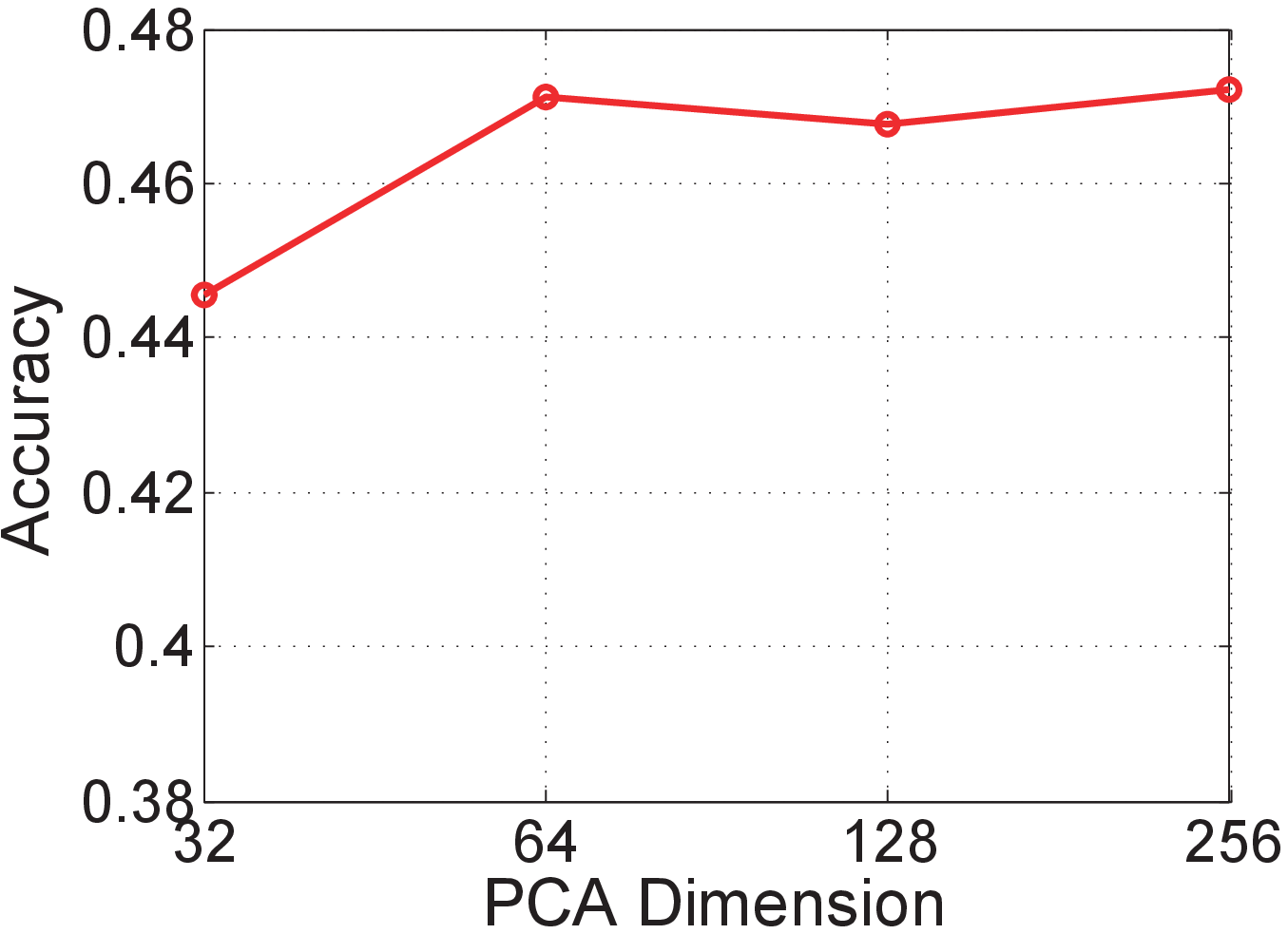}
  \includegraphics[width=0.48\linewidth]{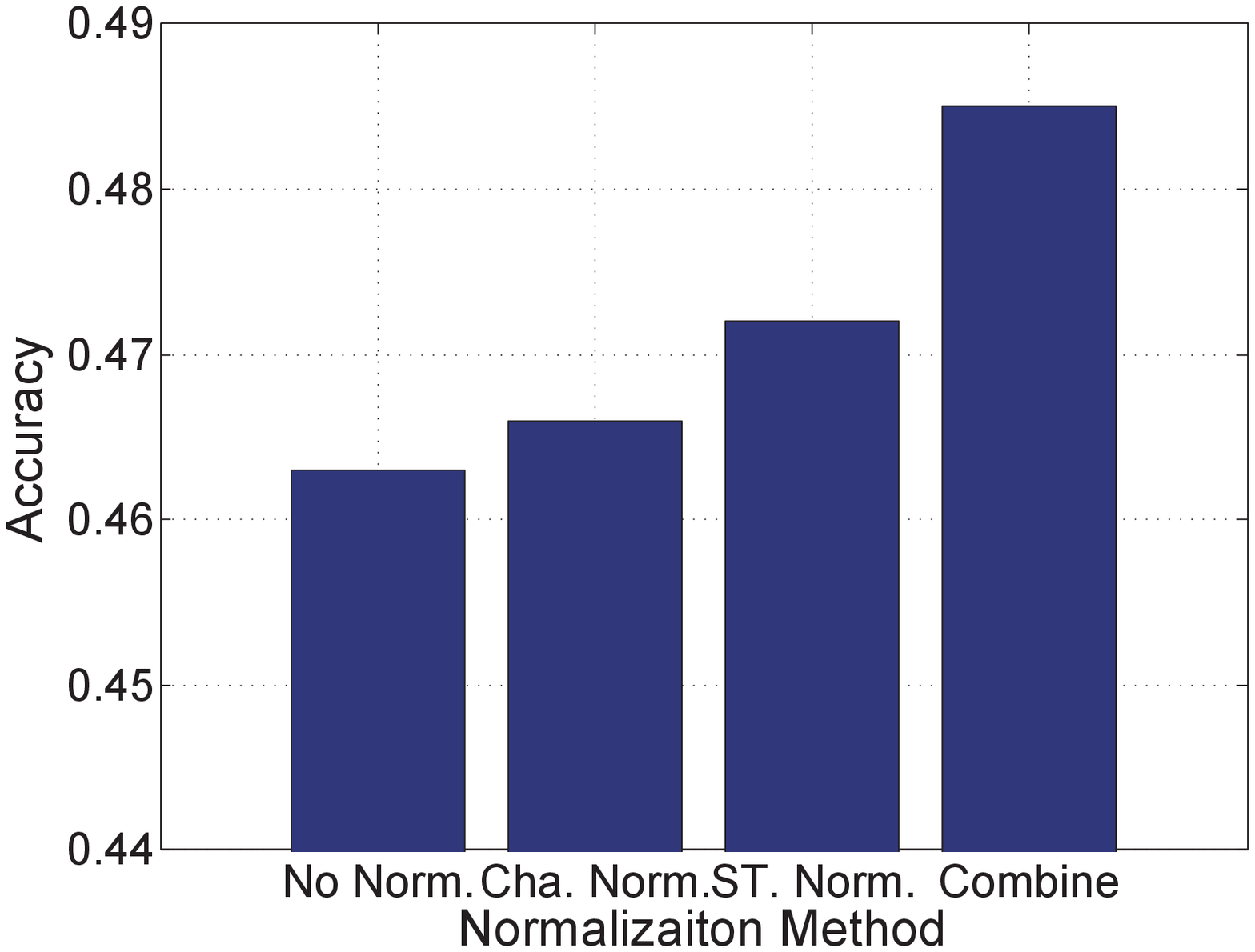}
  \caption{Exploration of different settings in TDD on the HMDB51 dataset. Left: Performance trend with varying PCA reduced dimension. Right: Comparison of different normalization methods. ``Combine'' means the fusion of spatiotemporal normalization and channel normalization.}
  \label{fig:pcanorm}
  \vspace{-3mm}
\end{figure}

\begin{table*}[t]
\centering
\begin{tabular}{|l|c|c|c|c|c|c|c|c|c|c|}
  \hline
  & \multicolumn{5}{|c|}{Spatial ConvNets} & \multicolumn{5}{|c|}{Temporal ConvNets} \\
  \hline
  Convolutional layer & conv1 & conv2 & conv3 & conv4 & conv5 & conv1  & conv2 & conv3 & conv4 & conv5\\
  \hline
  Recognition accuracy & 24.1\% & 33.9\% & 41.9\% & \textbf{48.5\%} & \textbf{47.2\%} & 39.2\% & 50.7\% & \textbf{54.5\%} & \textbf{51.2\%} & 46.1\%\\
  \hline
\end{tabular}
\vspace{1mm}
\caption{The performance of different layers of spatial nets and temporal nets on the HMDB51 dataset.}
\vspace{-3mm}
\label{tbl:layer}
\end{table*}

\begin{figure*}[t]
\center
  \subfigure[RGB]{
  \begin{minipage}[b]{0.132\linewidth}
    \includegraphics[width=\linewidth]{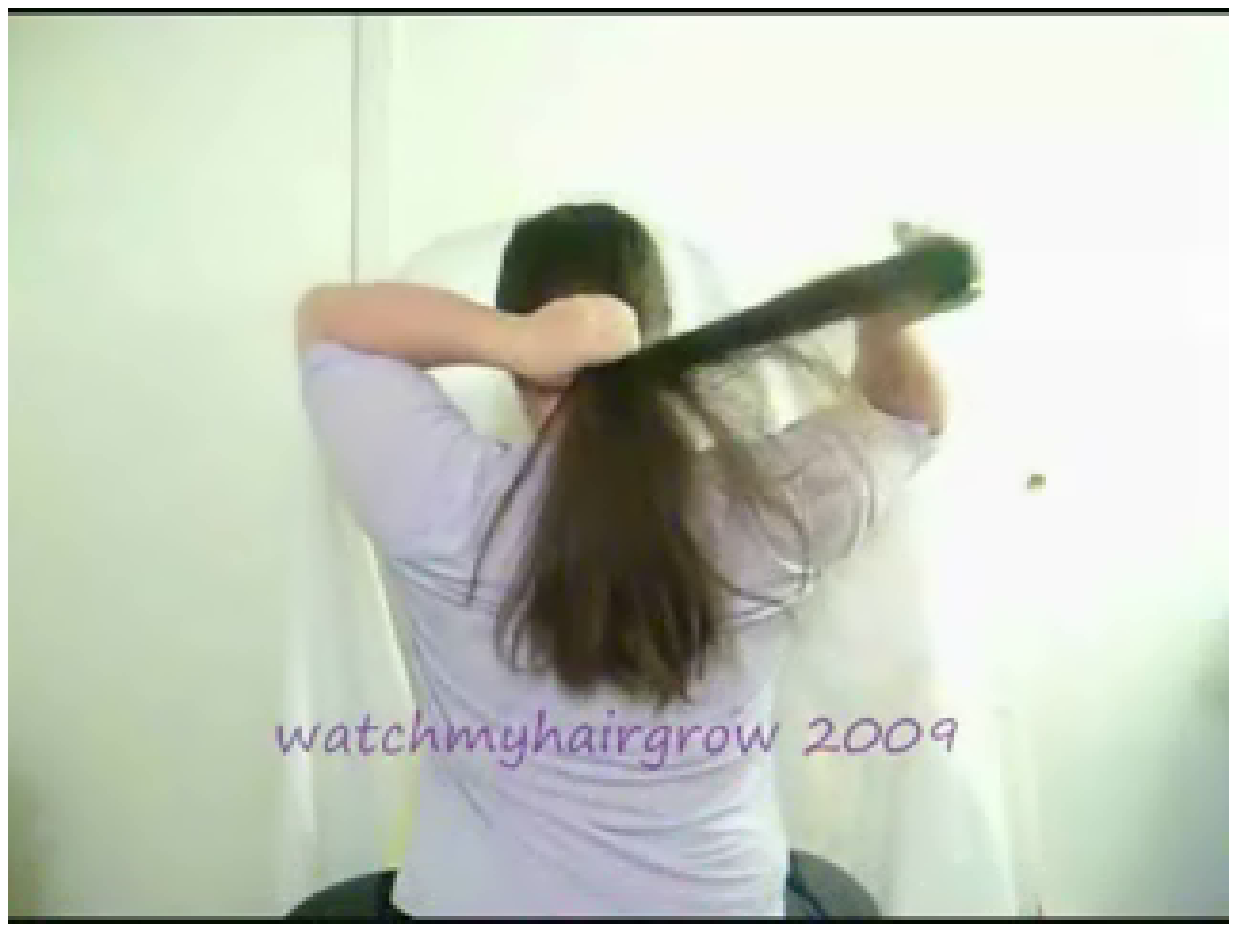}
    \includegraphics[width=\linewidth]{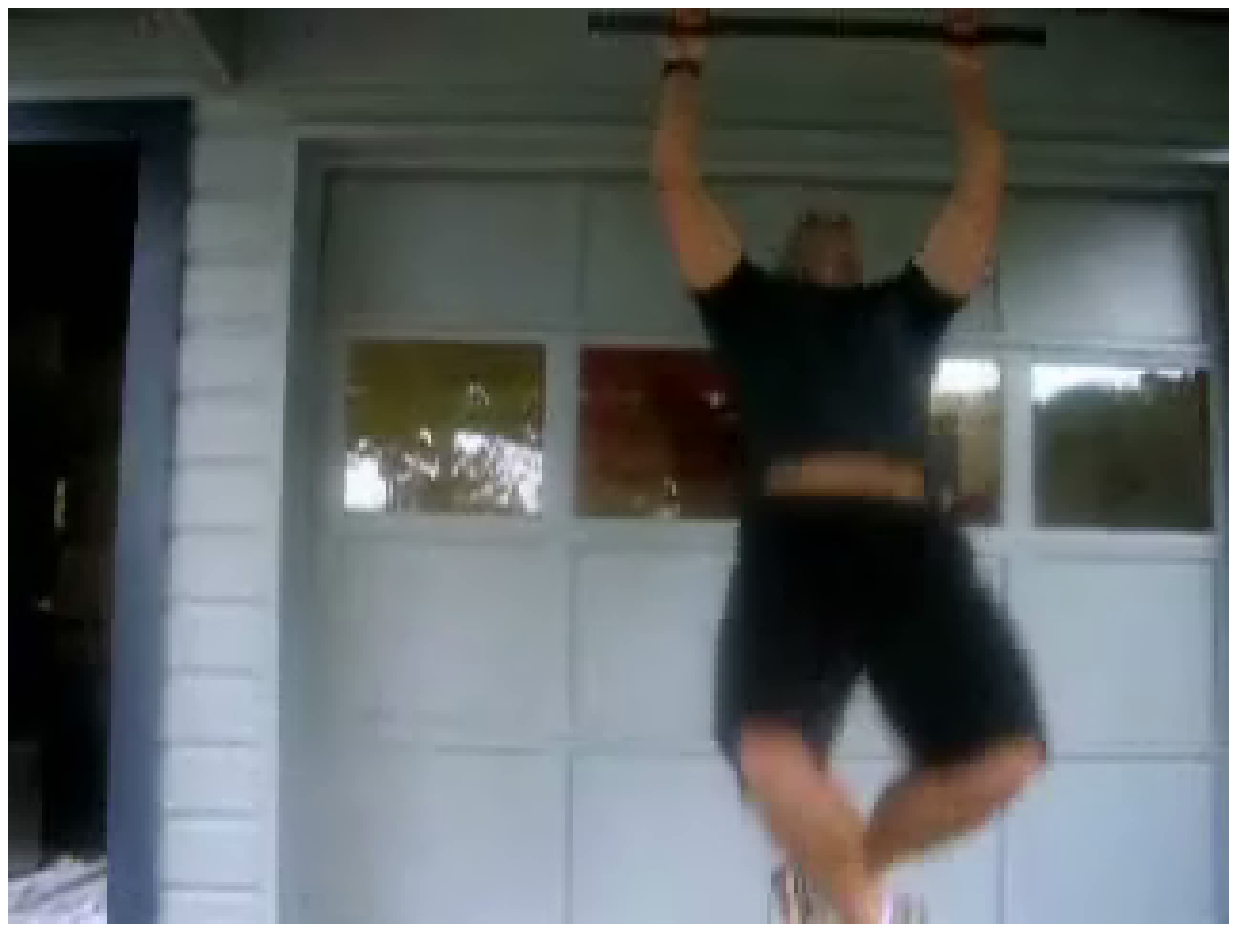}
    \includegraphics[width=\linewidth]{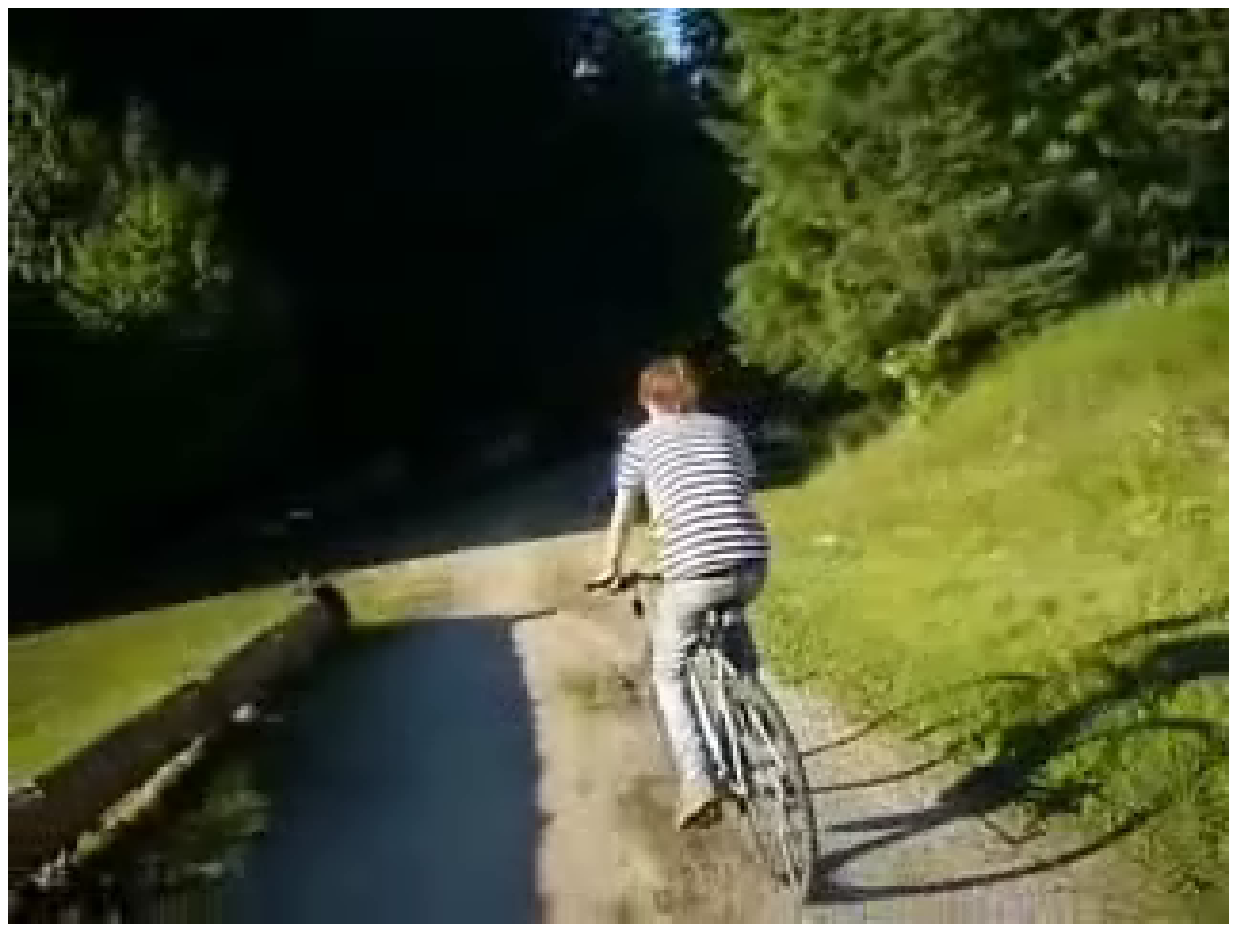}
  \end{minipage}
  }
  \hspace{-3mm}
  \subfigure[Flow-x]{
  \begin{minipage}[b]{0.1315\linewidth}
    \includegraphics[width=\linewidth]{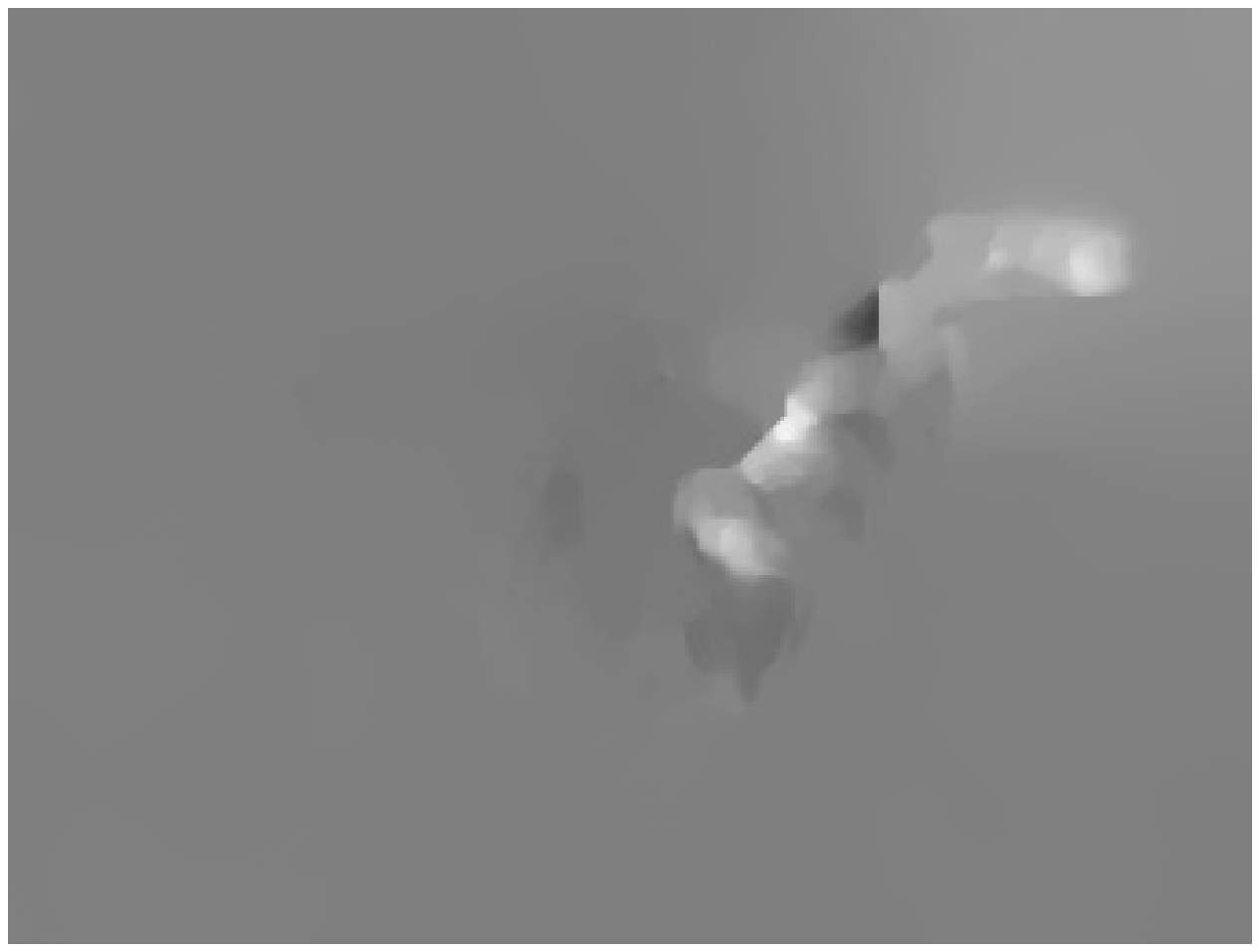}
    \includegraphics[width=\linewidth]{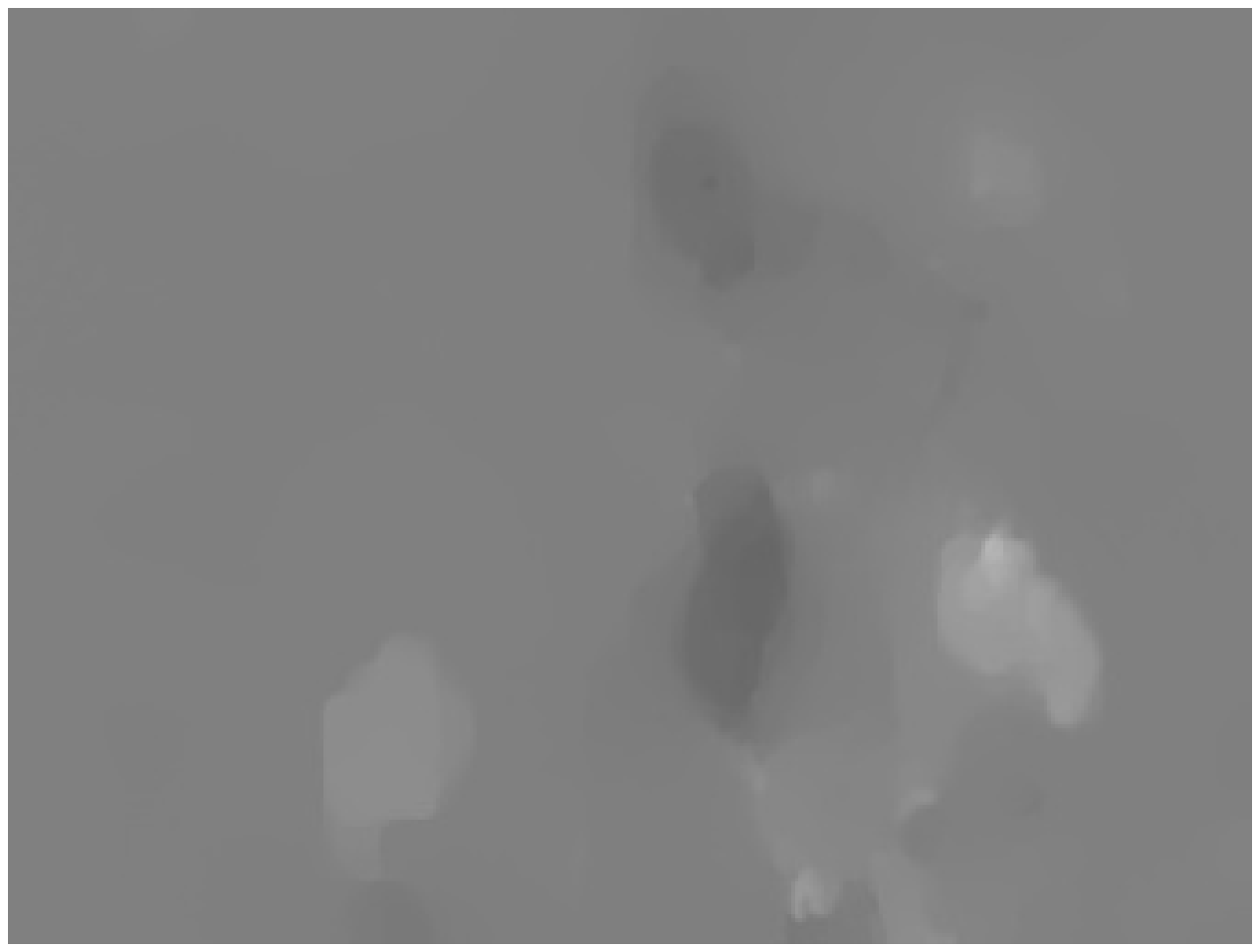}
    \includegraphics[width=\linewidth]{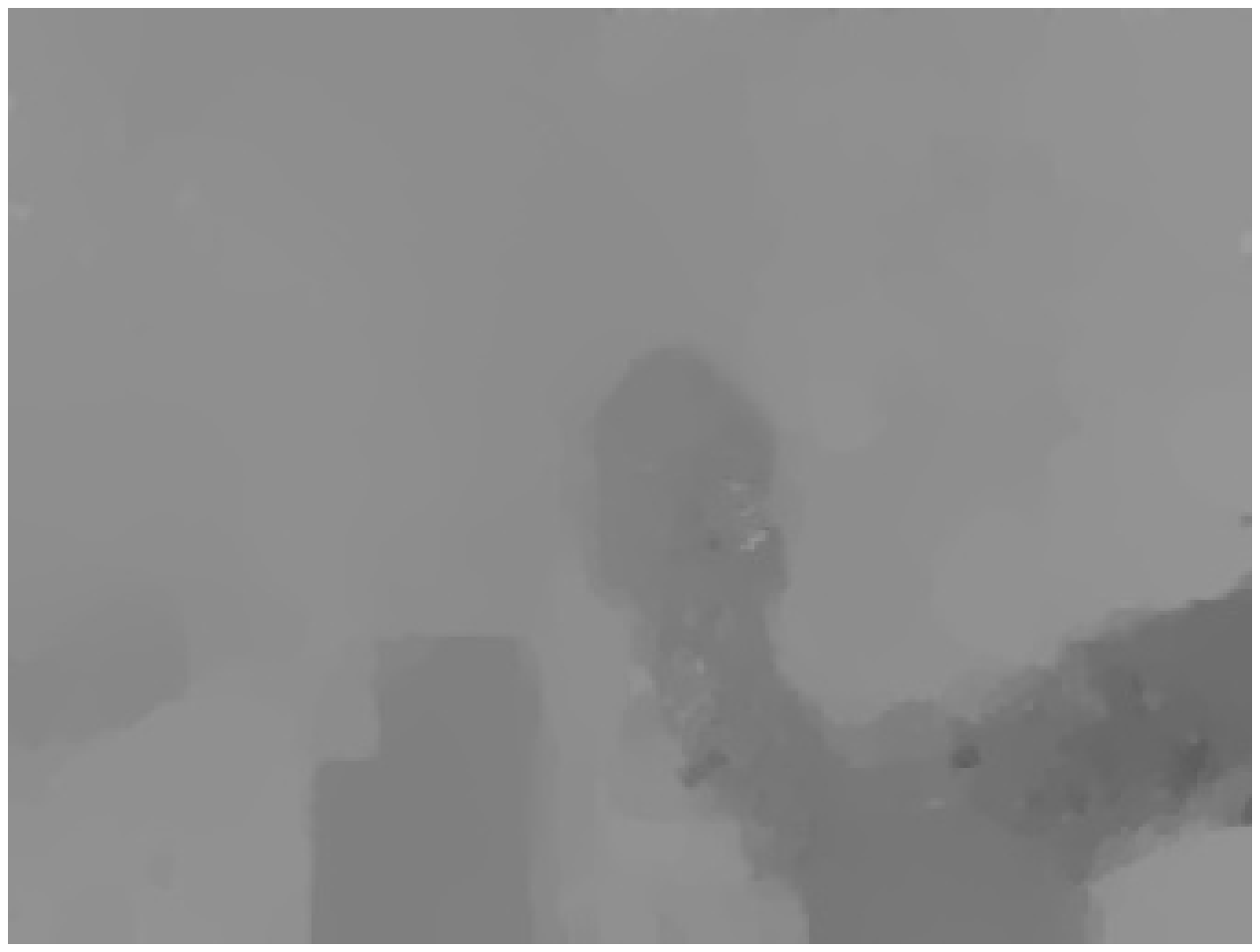}
  \end{minipage}
  }
  \hspace{-3mm}
  \subfigure[Flow-y]{
  \begin{minipage}[b]{0.1315\linewidth}
    \includegraphics[width=\linewidth]{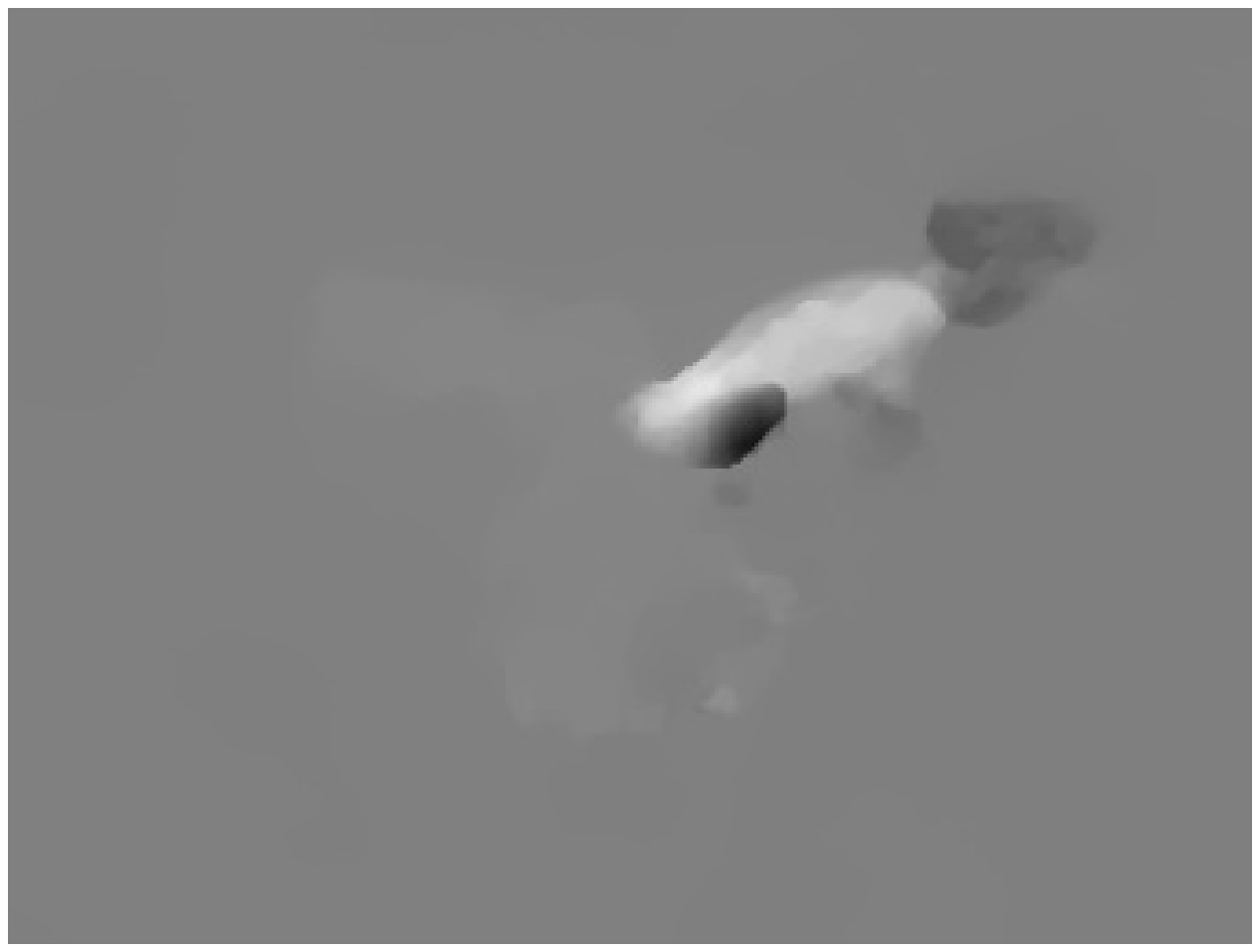}
    \includegraphics[width=\linewidth]{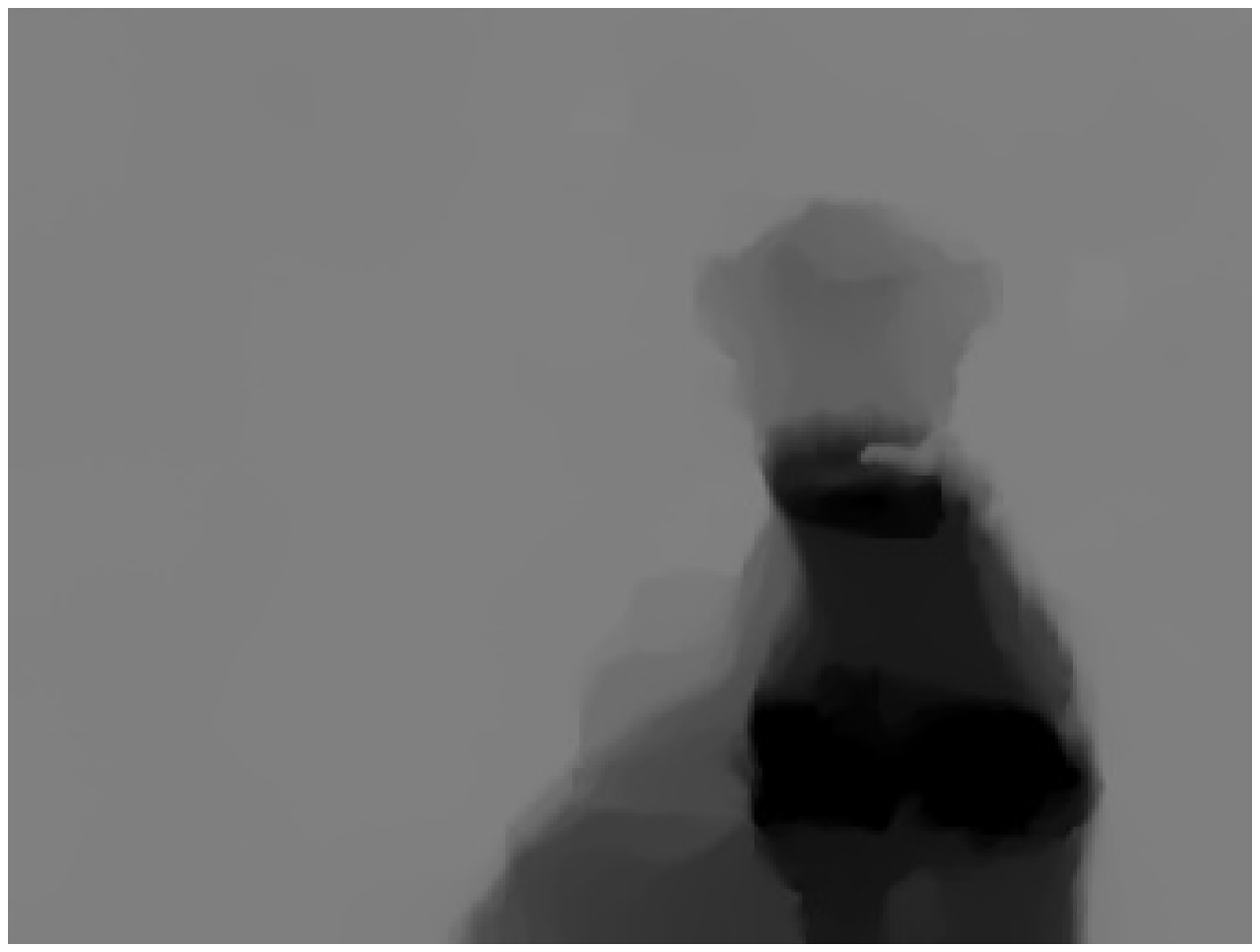}
    \includegraphics[width=\linewidth]{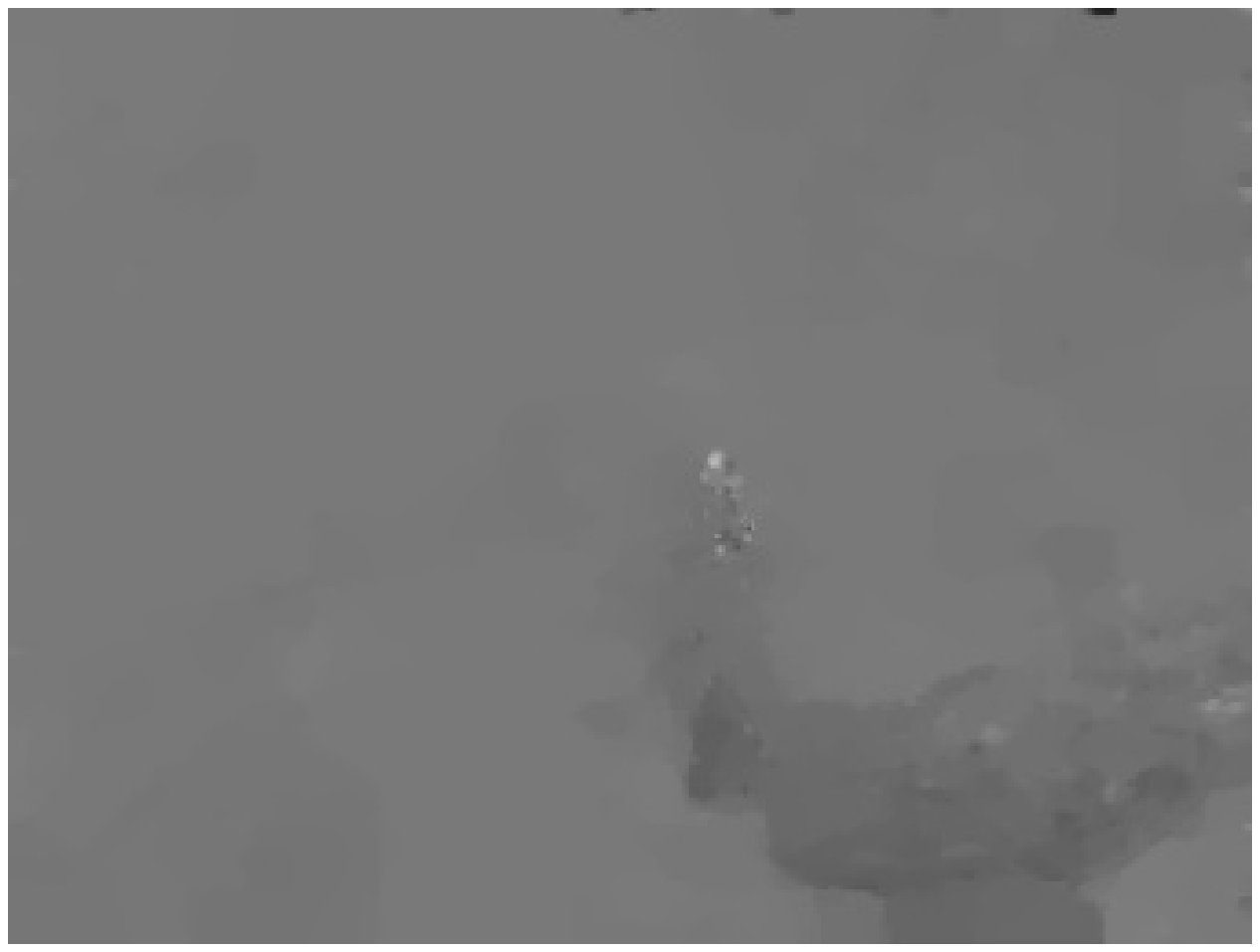}
  \end{minipage}
  }
  \hspace{-3mm}
  \subfigure[S-conv4]{
  \begin{minipage}[b]{0.132\linewidth}
    \includegraphics[width=\linewidth]{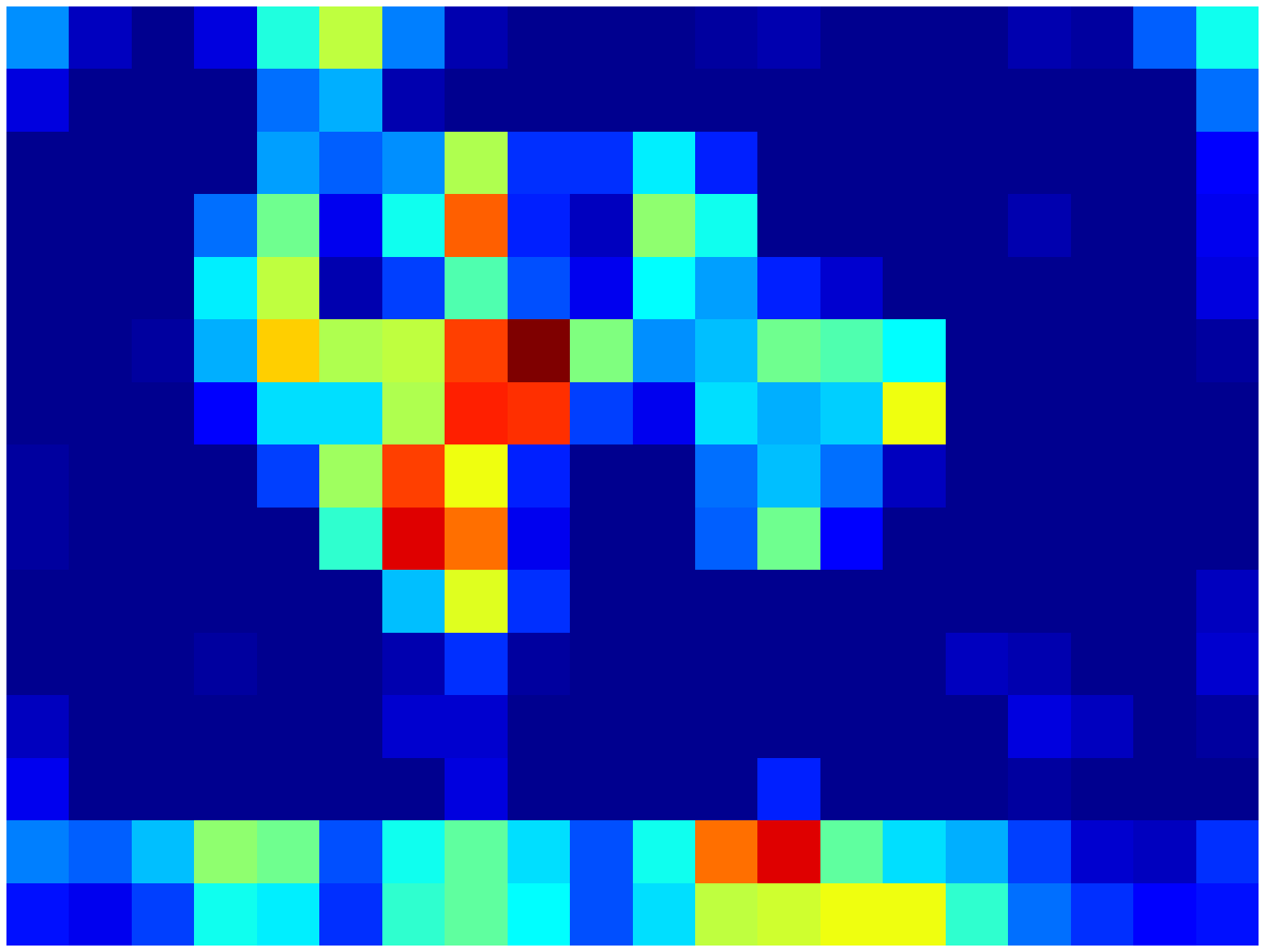}
    \includegraphics[width=\linewidth]{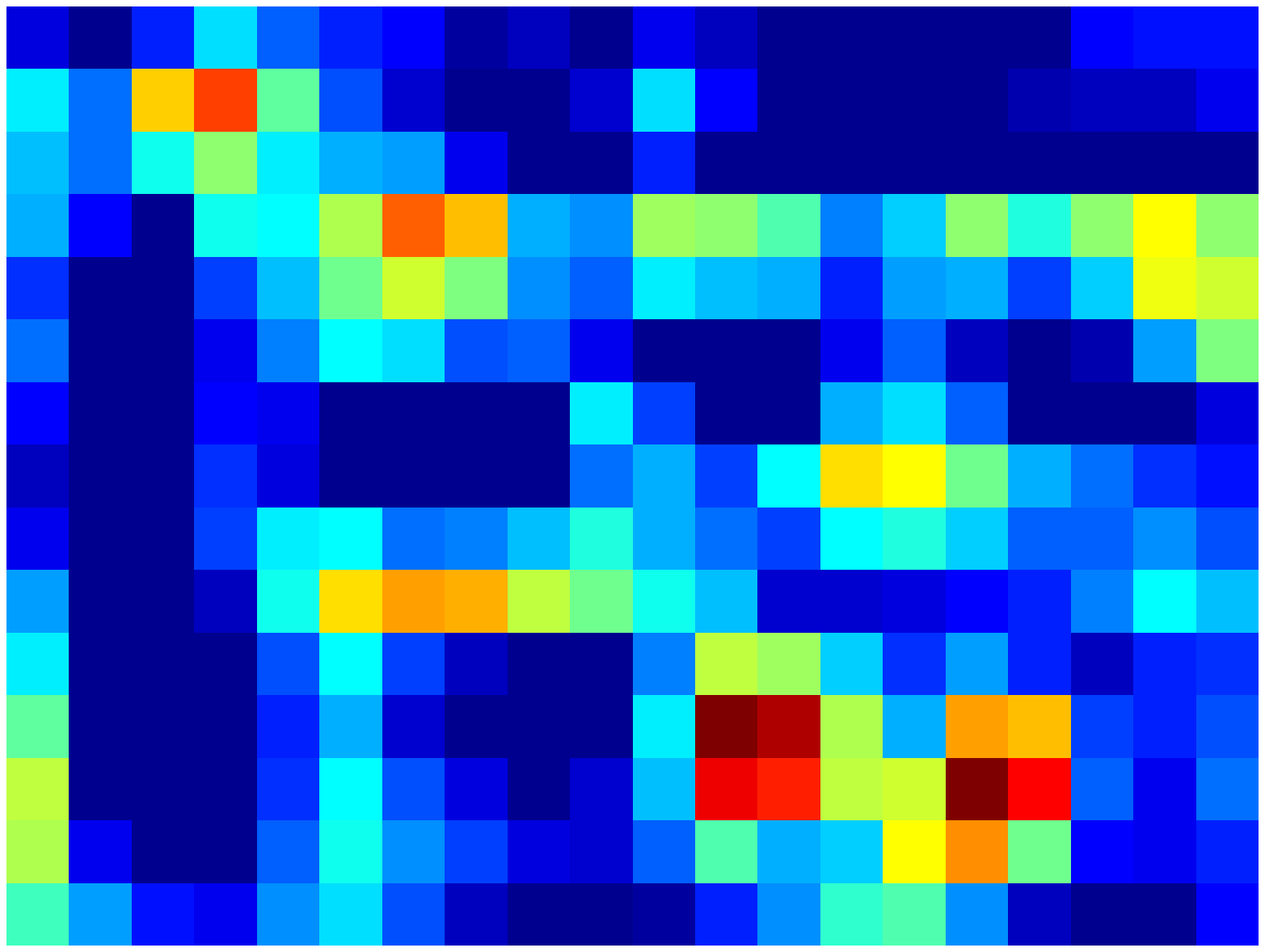}
    \includegraphics[width=\linewidth]{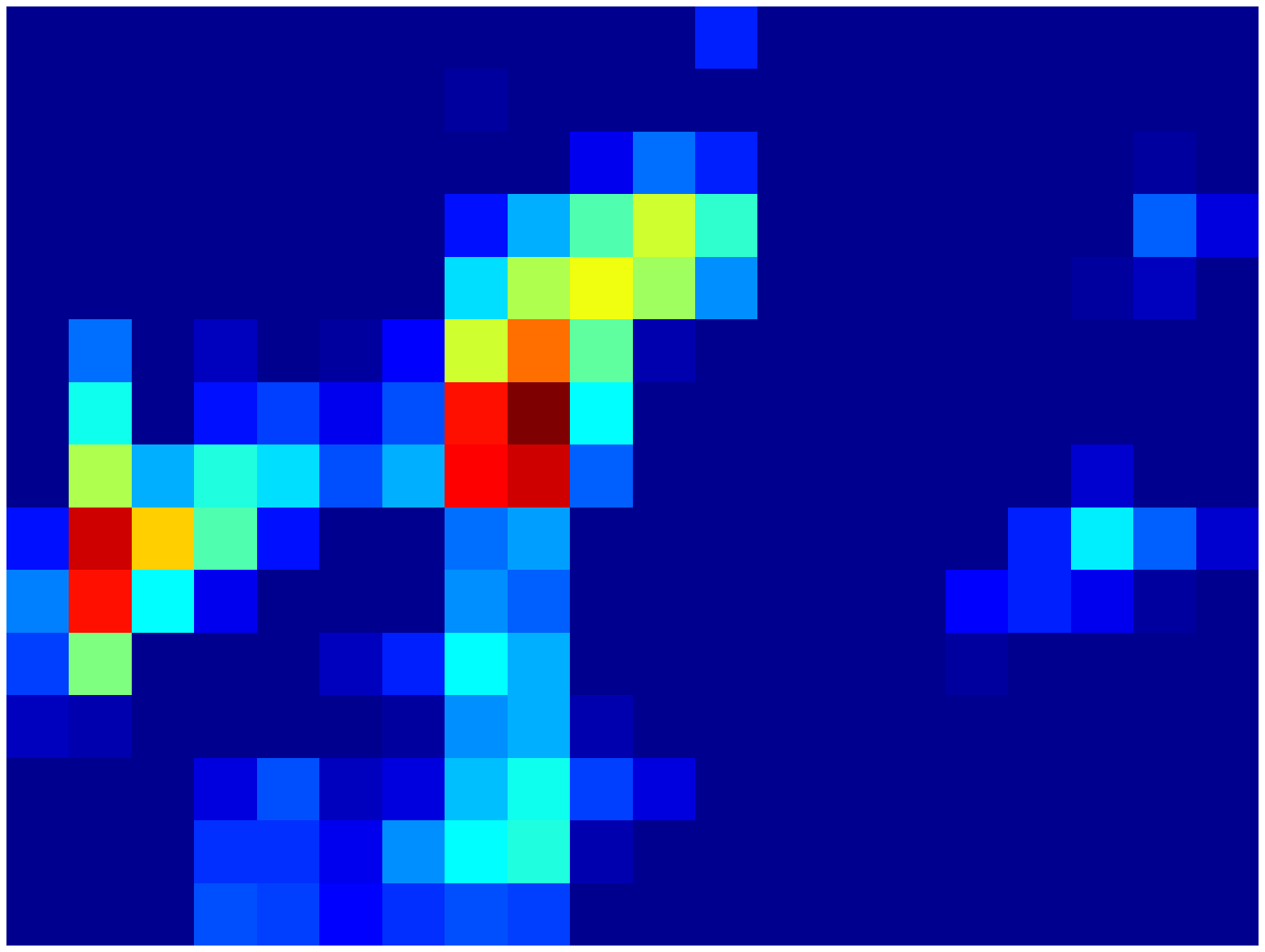}
  \end{minipage}
  }
  \hspace{-3mm}
  \subfigure[S-conv5]{
  \begin{minipage}[b]{0.132\linewidth}
    \includegraphics[width=\linewidth]{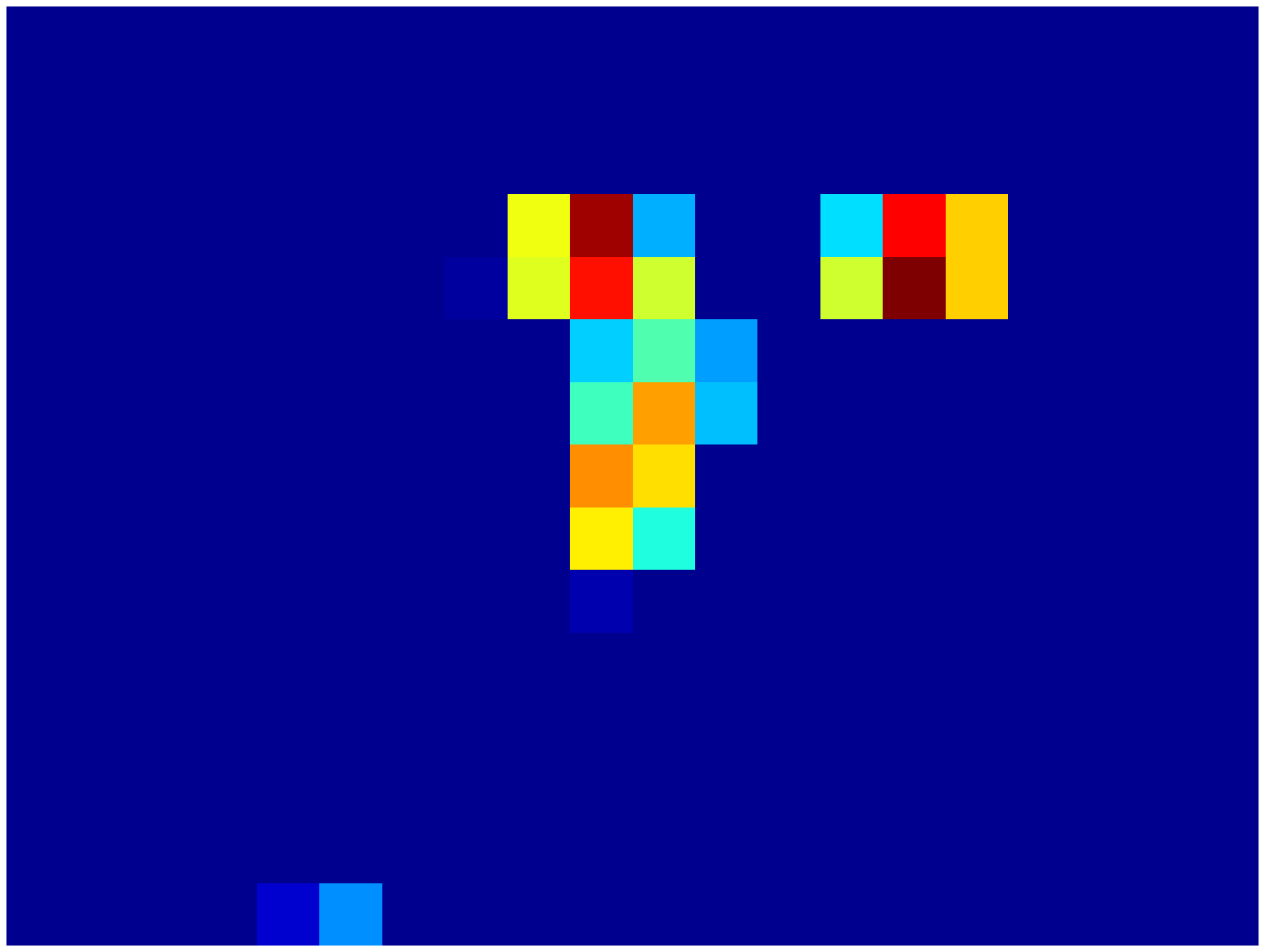}
    \includegraphics[width=\linewidth]{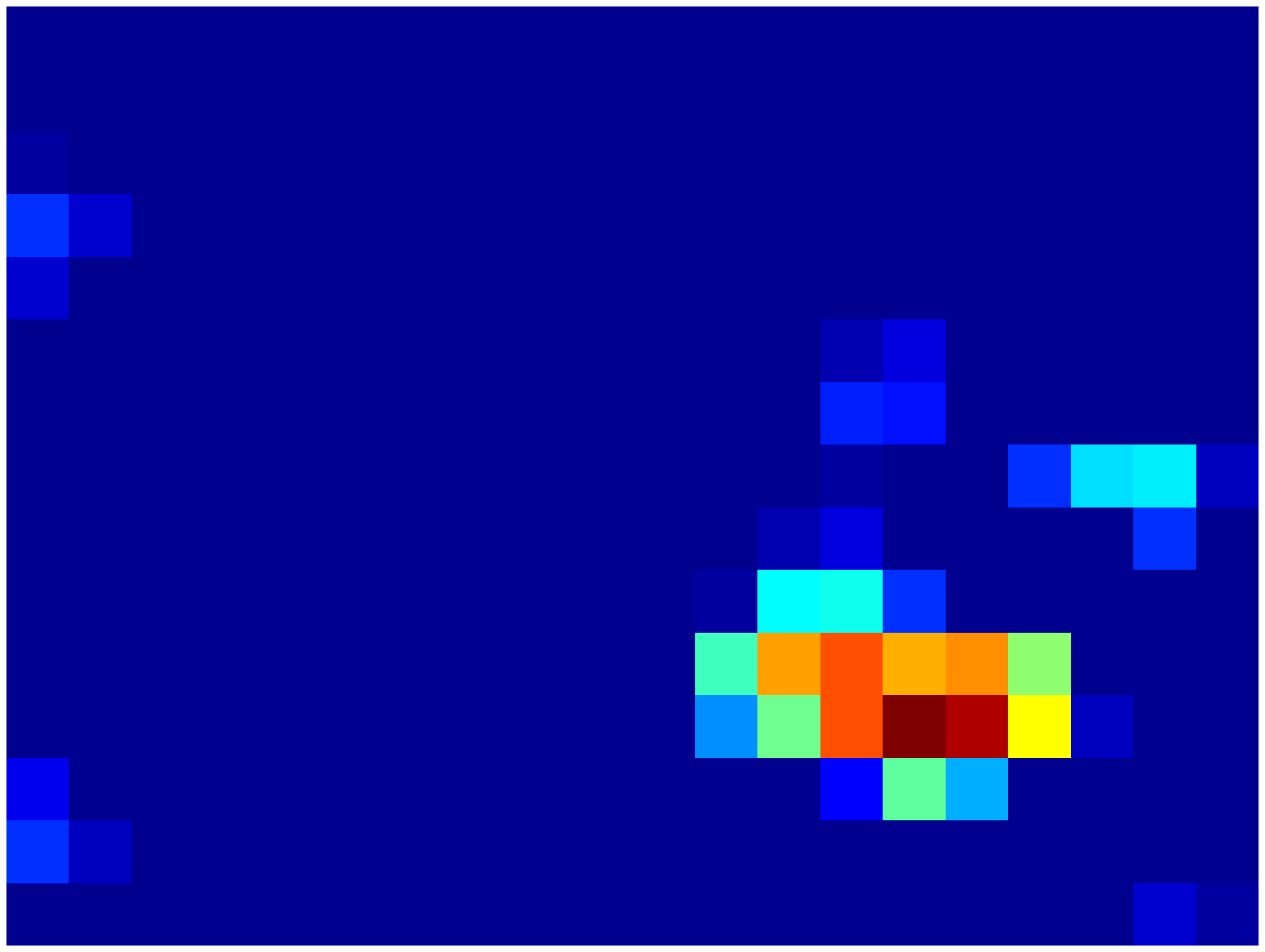}
    \includegraphics[width=\linewidth]{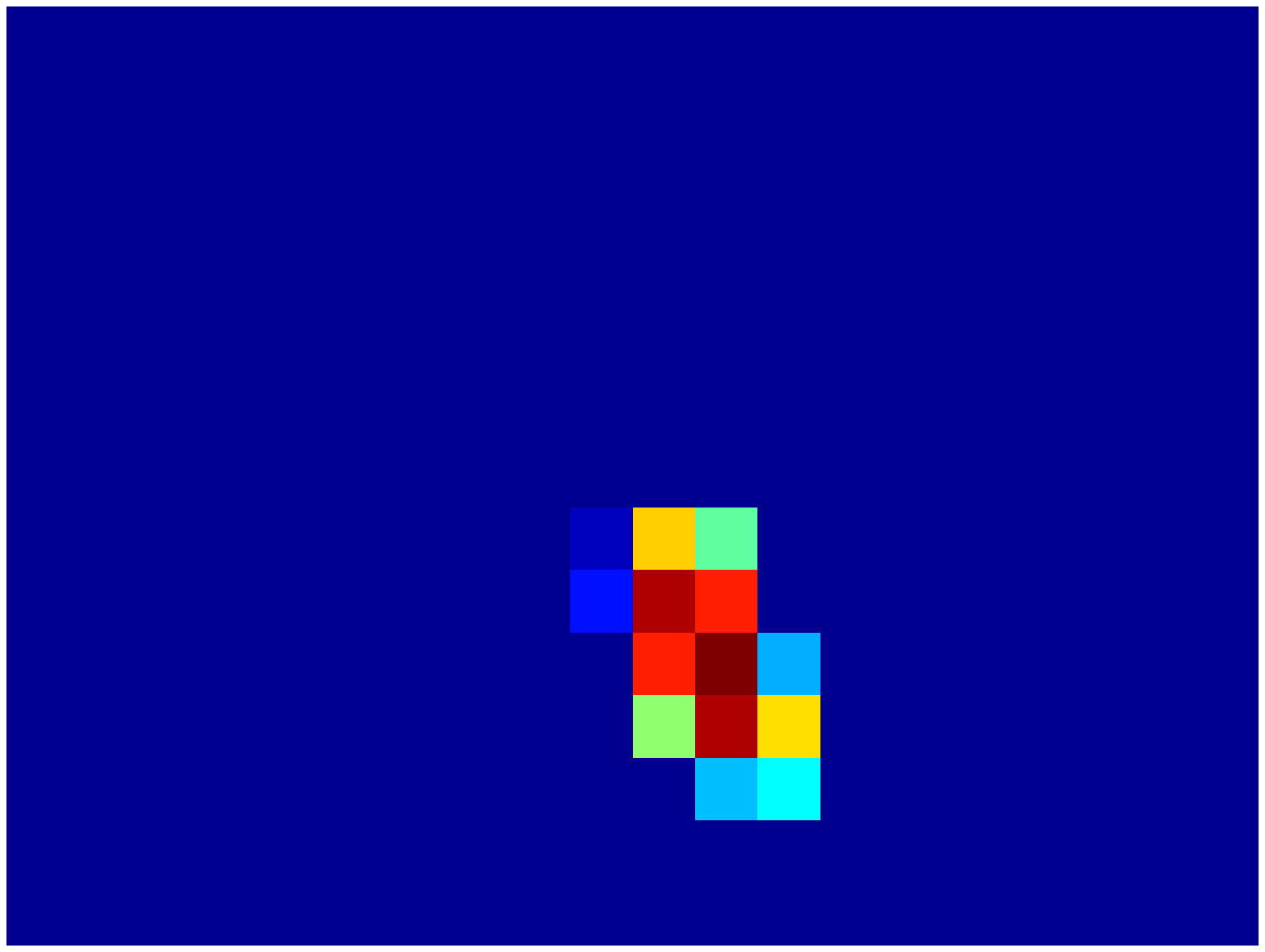}
  \end{minipage}
  }
  \hspace{-3mm}
    \subfigure[T-conv3]{
  \begin{minipage}[b]{0.132\linewidth}
    \includegraphics[width=\linewidth]{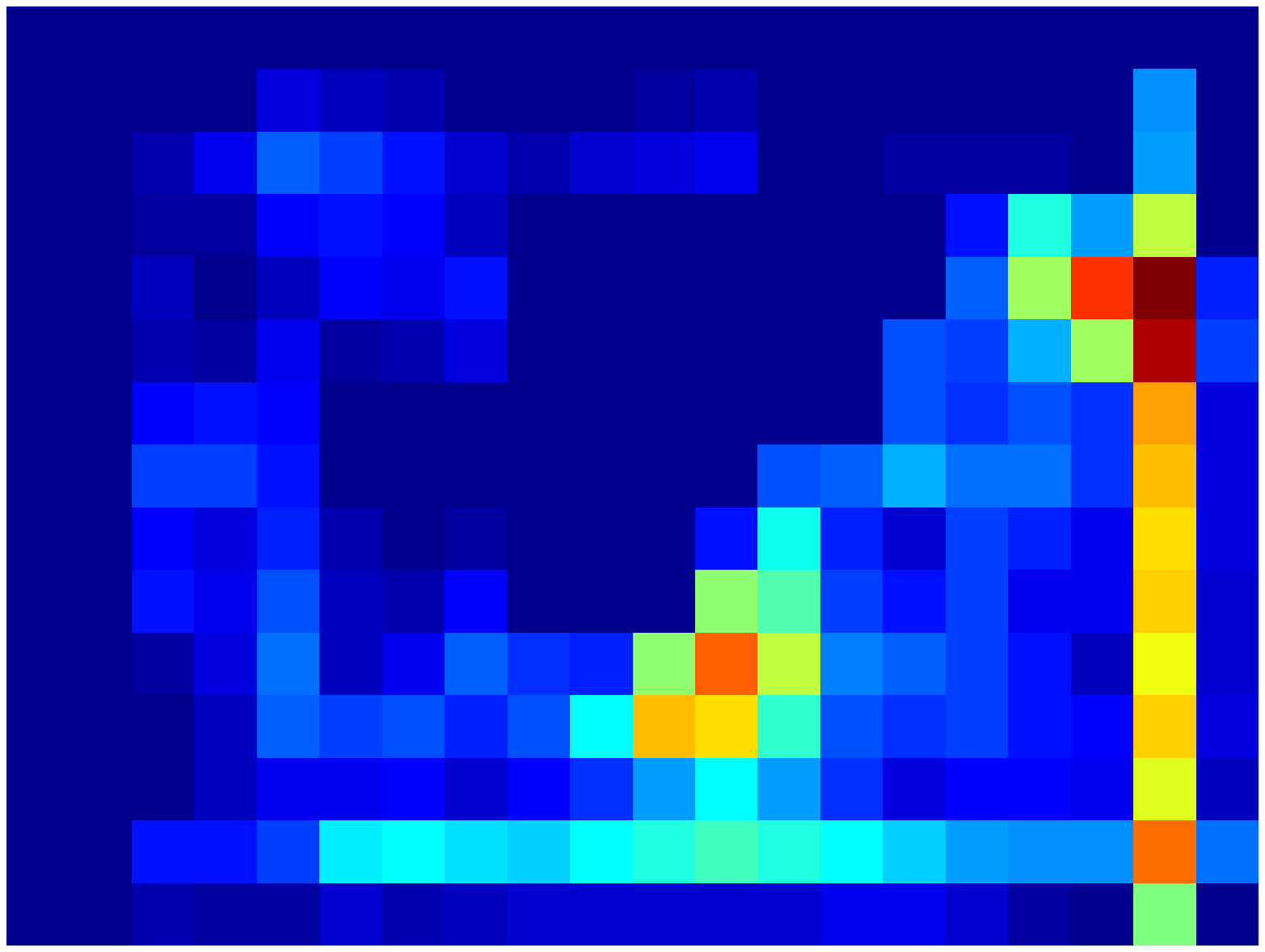}
    \includegraphics[width=\linewidth]{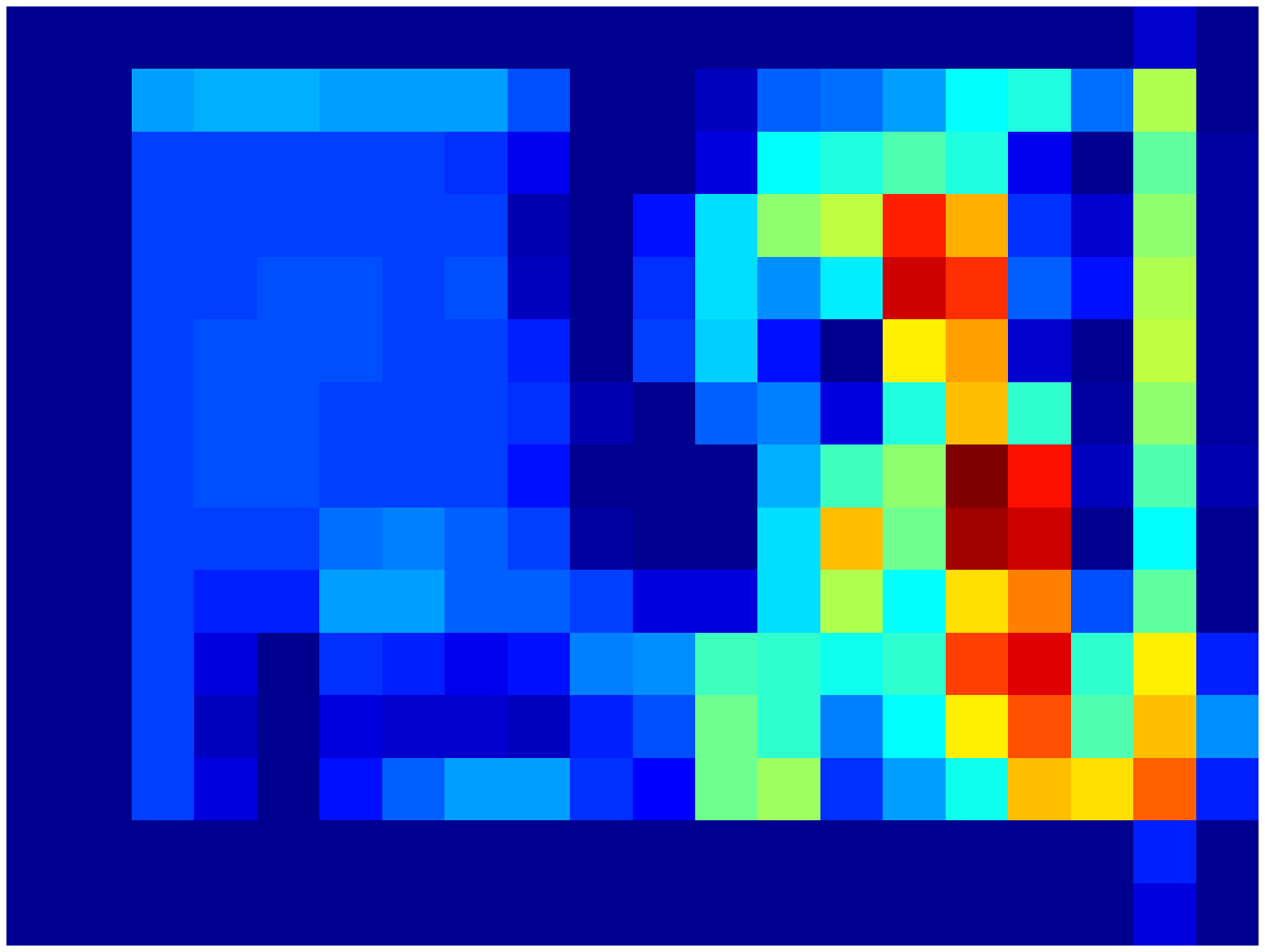}
    \includegraphics[width=\linewidth]{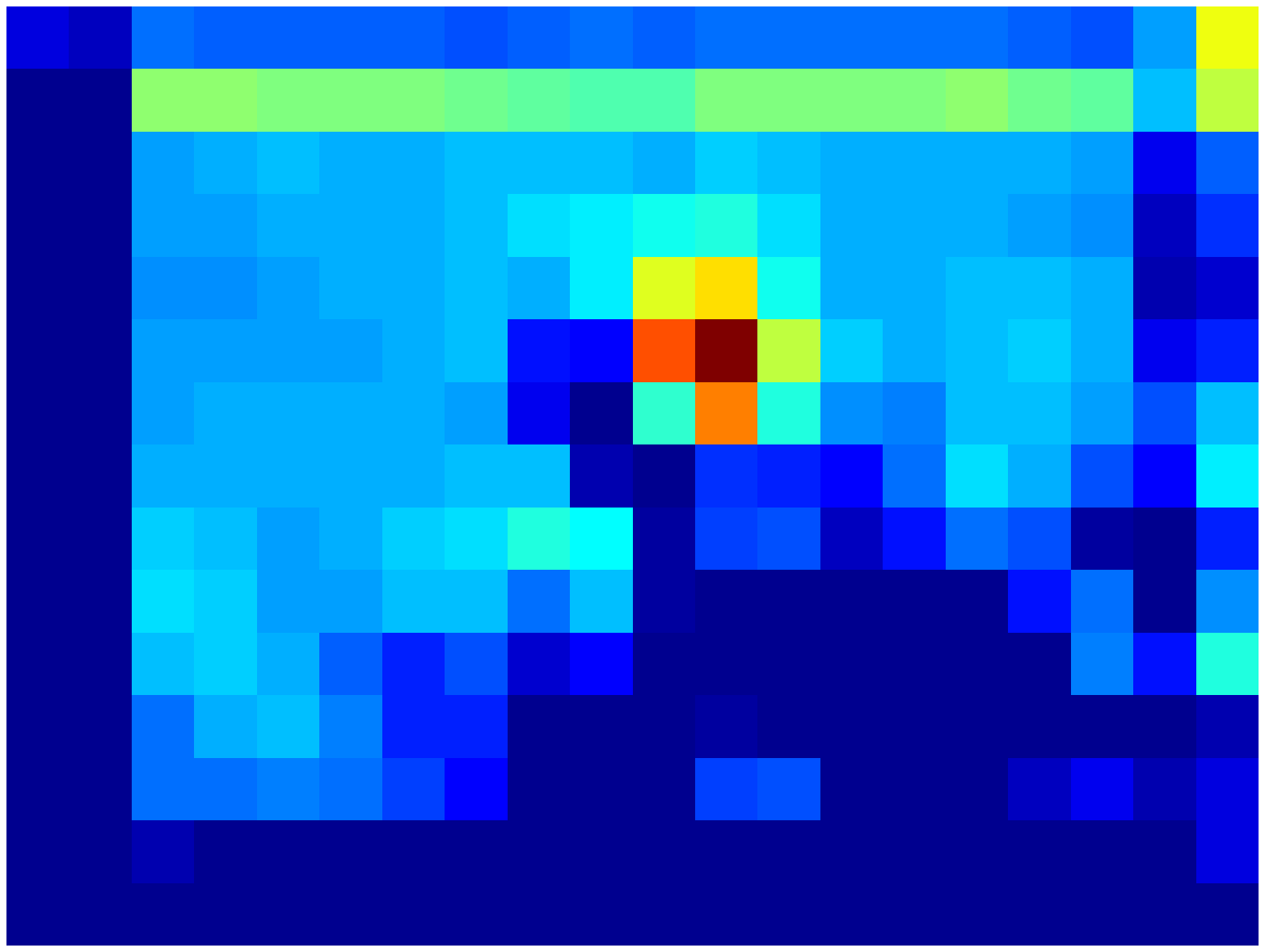}
  \end{minipage}
  }
  \hspace{-3mm}
    \subfigure[T-conv4]{
  \begin{minipage}[b]{0.132\linewidth}
    \includegraphics[width=\linewidth]{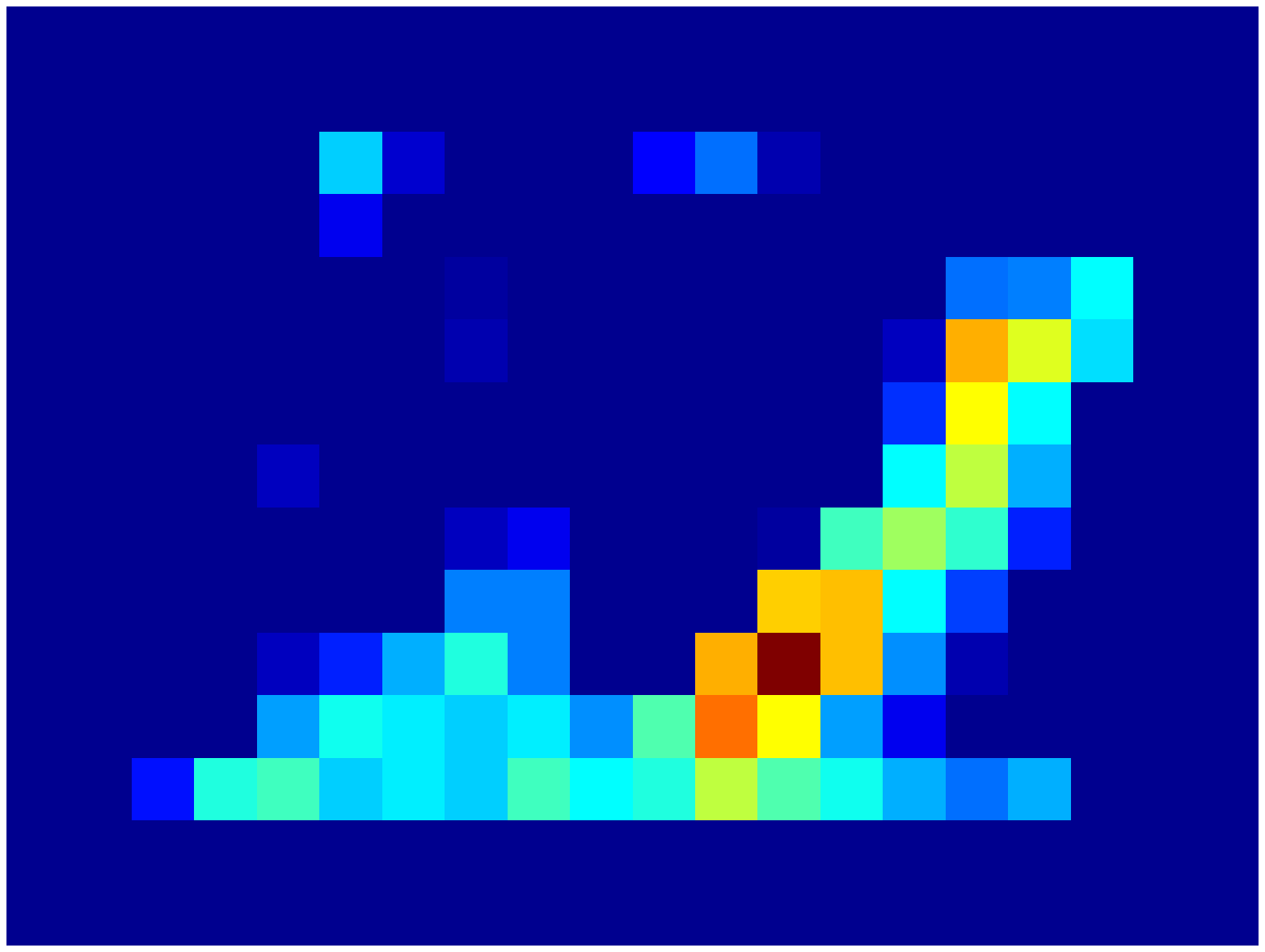}
    \includegraphics[width=\linewidth]{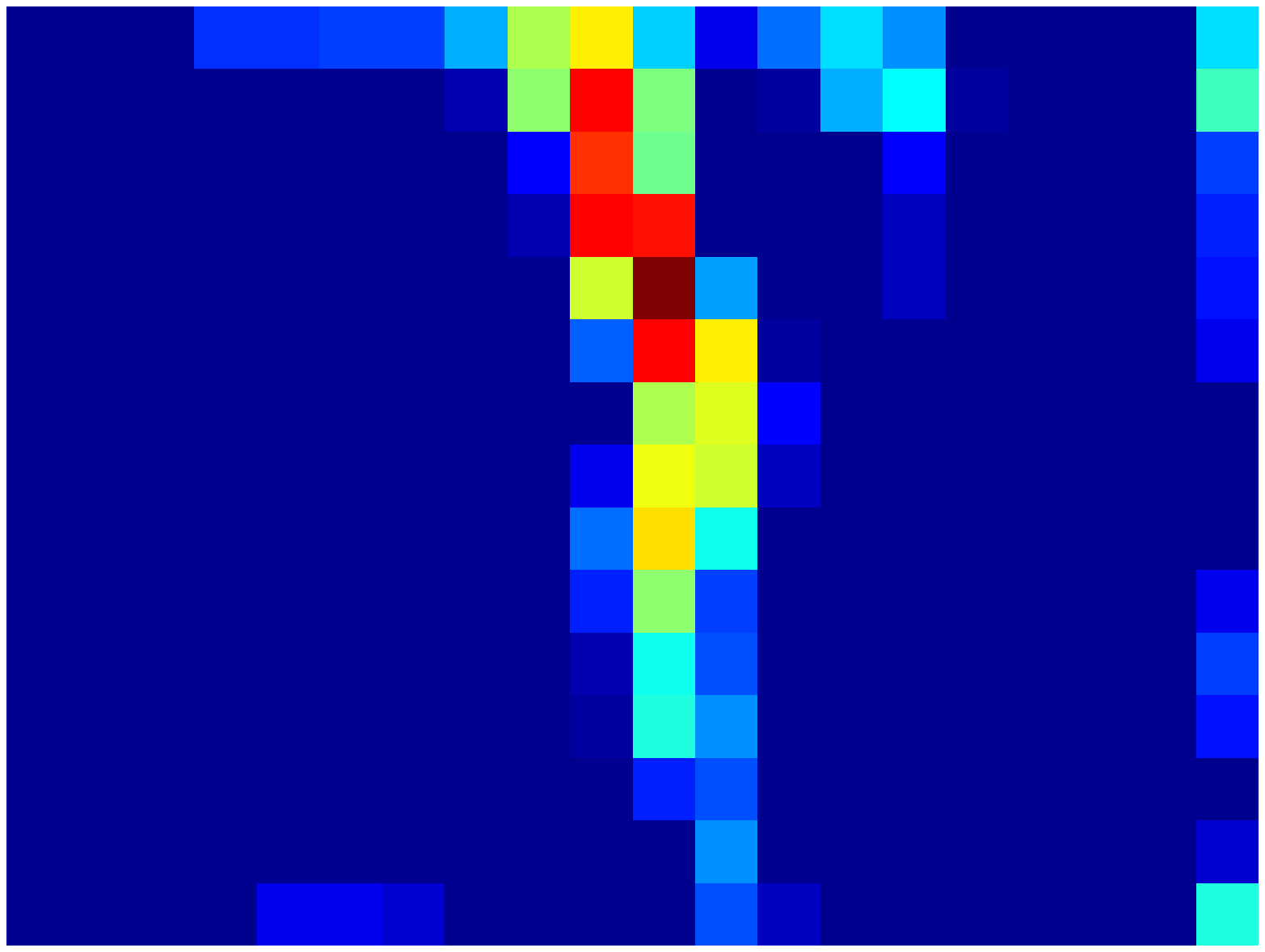}
    \includegraphics[width=\linewidth]{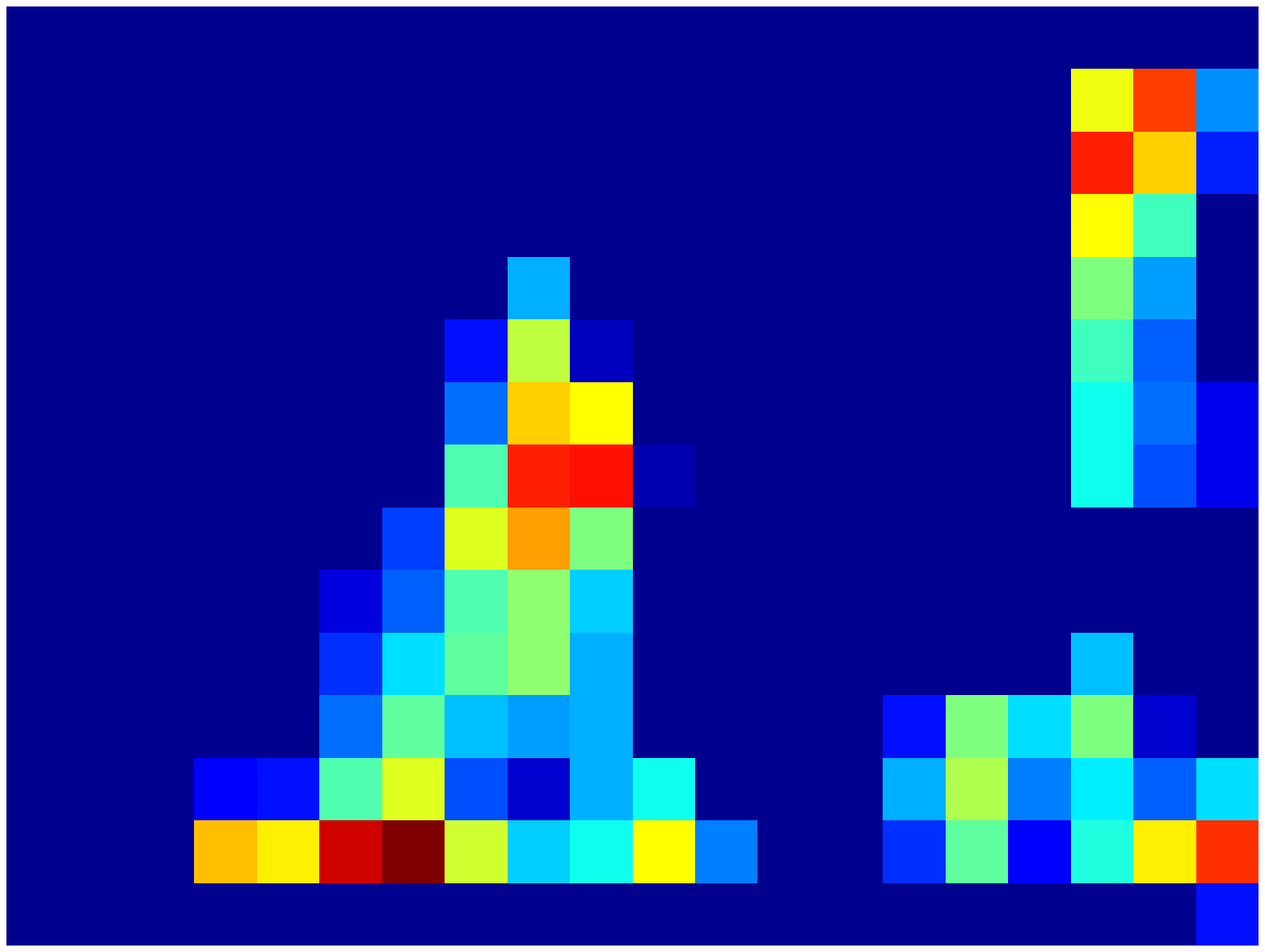}
  \end{minipage}
  }
  \caption{Examples of video frames, optical flow fields, and their corresponding feature maps of spatial nets and temporal nets.}
  \vspace{-0.5cm}
  \label{fig:maps}
\end{figure*}

\textbf{Dimension reduction.} To specify the PCA dimension of TDD for GMM training and Fisher vector encoding, we first explore different dimensions reduced by PCA on the HMDB51 dataset, with conv4 descriptors from spatial net. In this exploration experiment, we use the spatiotemporal normalization method for TDD and the results are shown in the left of Figure \ref{fig:pcanorm}. We vary the dimension from 32 to 256 and the results show that dimension 64 achieves the high performance, and higher dimension may cause performance degradation. Thus, we fix the dimension as $64$ for TDDs in the remainder of this section.

\textbf{Normalization method.} Another important component in TDD design is the normalization method and we have presented two normalization methods: spatiotemporal normalization (ST. Norm.) and channel normalization (Cha. Norm.) in Section \ref{sec:TDD} . We conduct experiments to investigate the effectiveness of normalization methods by using conv4 descriptors from spatial net on the HMDB51 dataset, and the results are shown in the right of Figure \ref{fig:pcanorm}. We see that normalization is important for improving performance and spatiotemporal normalization is the best choice. We also explore the complementary property of these two normalization methods by fusing the Fisher vectors of them, and observe that it can further improve the performance. Therefore, in the remainder of this section, we will use the combined representation obtained from these two normalization methods for TDDs.

\textbf{Different layers.} Finally we investigate the performance of TDDs from different layers of spatial and temporal nets on the HMDB51 dataset, and the results are summarized in Table \ref{tbl:layer}. For layers of conv5, conv4, and conv3, we use the outputs of RELU activations, and for layers of conv2 and conv1, we choose the outputs of max pooling layers after convolution operations. We see that descriptors of layers conv4 and conv5 obtain highest recognition performance for spatial net, while the ones of layers conv3 and conv4 are top performers for temporal net. Therefore, in the following evaluation of TDD, we choose the descriptors from conv4 and conv5 layers for spatial nets, and conv3 and conv4 layers for temporal nets.

\subsection{Evaluation of TDDs}
\begin{table}
\small
\begin{center}
\begin{tabular}{|l|c|c|}
  \hline
  Algorithm & HMDB51  & UCF101 \\
  \hline
  \hline
  HOG \cite{WangS13a,WangS13b} &  40.2\% &  72.4\% \\
  \hline
  \hline
  HOF \cite{WangS13a,WangS13b} & 48.9\% &  76.0\% \\
  \hline
  MBH \cite{WangS13a,WangS13b} & 52.1\%  &  80.8\% \\
  \hline
  HOF+MBH \cite{WangS13a,WangS13b} & 54.7\% & 82.2\% \\
  \hline
  \hline
  iDT \cite{WangS13a,WangS13b} & \textbf{57.2\%} & \textbf{84.7\%} \\
  \hline
  \hline
  Spatial net \cite{SimonyanZ14} & 40.5\% & 73.0\% \\
  \hline
  Temporal net \cite{SimonyanZ14} & 54.6\% & 83.7\% \\
  \hline
  Two-stream ConvNets \cite{SimonyanZ14} & \textbf{59.4\%} & \textbf{88.0\%} \\
  \hline
  \hline
  Spatial conv4 & 48.5\% &  81.9\% \\
  \hline
  Spatial conv5 & 47.2\% & 80.9\% \\
  \hline
  Spatial conv4 and conv5 & \textbf{50.0\%} & \textbf{82.8\%} \\
  \hline
  \hline
  Temporal conv3 &  54.5\% &  81.7\% \\
  \hline
  Temporal conv4 & 51.2\% &  80.1\% \\
  \hline
  Temporal conv3 and conv4 & \textbf{54.9\%} & \textbf{82.2\%}\\
  \hline
  \hline
  TDD & 63.2\% & 90.3\% \\
  \hline
  TDD and iDT & \textbf{65.9\%} & \textbf{91.5\%}\\
  \hline
\end{tabular}
\vspace{1mm}
\caption{Performance of TDD on the HMDB51 dataset and UCF101 dataset. We compare our proposed TDD with iDT features \cite{WangS13a} and two-stream ConvNets \cite{SimonyanZ14}. We also explore the complementary properties TDD features and iDT features. The combination of them can further boost the performance.}
\vspace{-7mm}
\label{tbl:TDD}
\end{center}
\end{table}

In this section, we evaluate the performance of our proposed TDDs on the HMDB51 and UCF101 dataset, and the experimental results are summarized in Table \ref{tbl:TDD}. We first compare the performance of TDDs with that of improved trajectories. The convolutional descriptors of spatial net are much better than HOG descriptors, which indicates that deep-learned features contains more discriminative capacity than hand-crafted features. For convolutional descriptors of temporal net, they are better than or comparable to the descriptors of HOF and MBH, but the improvement is not so evident as spatial convolutional descriptors. The reason may be that HOF and MBH calculation is based on warped optical flow instead of original optical flow, which has been proved to be pretty effective for HOF descriptor \cite{WangS13a}. We consider using warped flow for TDDs extraction in the future.

We also compare the performance of TDDs with the two-stream ConvNets. Although our trained two-stream ConvNets obtain slightly lower performance than theirs, we see that our spatial TDDs outperform spatial nets by a large margin and temporal TDD is comparable to their temporal net. These results indicate the fact that trajectory-constrained sampling and pooling is an effective strategy for improving recognition performance, in particular for spatial TDDs. We also notice that the combined TDDs from spatial and temporal nets outperform two-stream ConvNets by around $4\%$ and $2\%$ on the two datasets, respectively. We also show some examples of video frames, optical flow fields, and their corresponding feature maps in Figure \ref{fig:maps}. From these examples, we see that the convolutional feature maps are relatively sparse and exhibit high correlation with the action areas.

Finally, we explore a practical way to improve the recognition performance of action recognition system by combining TDDs with iDTs, using early fusion of Fisher vector representation. The recognition results are shown in Table \ref{tbl:TDD}, and the fusion of them can further boost the performance. This further improvement indicates our TDDs are complementary to those low-level local features.

\textbf{Computational costs.} Compared with iDT, we only track points on a single scale and extract original flow instead of warped flow. The ConvNets are implemented by Cuda and computing is very efficient.

\subsection{Comparison to the state of the art}
\begin{table}
\small
\centering
\begin{tabular}{|lr|lr|}
\hline
\multicolumn{2}{|c|}{HMDB51} & \multicolumn{2}{|c|}{UCF101} \\
\hline
\hline
STIP+BoVW \cite{KuehneJGPS11} & 23.0\% & STIP+BoVW \cite{Soomro12} & 43.9\% \\
Motionlets \cite{WangQT13a} & 42.1\% & Deep Net \cite{KarpathyTSLSF14} & 63.3\% \\
DT+BoVW \cite{WangKSL13} & 46.6\% & DT+VLAD \cite{CaiWPQ14} & 79.9\% \\
DT+MVSV \cite{CaiWPQ14} & 55.9\% & DT+MVSV \cite{CaiWPQ14} & 83.5\% \\
iDT+FV \cite{WangS13a} & 57.2\% & iDT+FV \cite{WangS13b} & 85.9\% \\
iDT+HSV \cite{PengWWQ14} & 61.1\% & iDT+HSV \cite{PengWWQ14} & 87.9\% \\
Two Stream \cite{SimonyanZ14} & 59.4\% & Two Stream \cite{SimonyanZ14} & 88.0\% \\
\hline
\hline
TDD+FV & 63.2\% &TDD+FV & 90.3\%  \\
Our best result & \textbf{65.9\%} &Our best result & \textbf{91.5\%}\\
\hline
\end{tabular}
\vspace{1mm}
\caption{Comparison of TDD to the state of the art. We separately present the results of TDDs and our best results obtained with early fusion of TDDs and iDTs.}
\vspace{-5mm}
\label{tbl:stoa}
\end{table}

Table \ref{tbl:stoa} compares our recognition results with several recently published methods on the dataset of HMDB51 and UCF101. The performance of TDDs outperforms previous methods on both datasets. On the HMDB51 dataset, our best result outperforms other methods by $4.8\%$, and on the UCF101 dataset, our best result outperforms by $3.5\%$. This superior performance of TDDs indicates the effectiveness of introducing trajectory-constrained sampling and pooling into deep-learned features.

\section{Conclusions}
This paper has proposed an effective video presentation, called trajectory-pooled deep-convolutional descriptor (TDD), which integrates the advantages of hand-crafted and deep-learned features. Deep architectures are utilized to learn discriminative convolutional feature maps, and then the strategies of trajectory-constrained sampling and pooling are adopted to aggregate these convolutional features into TDDs. Our features achieve superior performance on two datasets for action recognition, as evidenced by comparison with the state-of-the-art methods.

\section*{Acknowledgement}
 This work is supported by a donation of Tesla K40 GPU from NVIDIA Corporation. Limin Wang is supported by Hong Kong PhD Fellowship. Yu Qiao is the corresponding author and supported by National Natural Science Foundation of China (91320101, 61472410), Shenzhen Basic Research Program (JCYJ20120903092050890, JCYJ20120617114614438, JCYJ20130402113127496), 100 Talents Program of CAS, and Guangdong Innovative Research Team Program (No.201001D0104648280).

{\small
\bibliographystyle{ieee}
\bibliography{reference}
}

\end{document}